%% file: main.tex
\documentclass[10pt,twocolumn,letterpaper]{article}

\usepackage[pagenumbers]{cvpr} 

\usepackage{graphicx}
\usepackage{amsmath}
\usepackage{amssymb}
\usepackage{booktabs}
\input{headers}
\usepackage{listings}
\usepackage{adjustbox}

\usepackage{algorithm}
\usepackage{algorithmicx}
\usepackage{algpseudocode}
\algdef{SE}[DOWHILE]{Do}{doWhile}{\algorithmicdo}[1]{\algorithmicwhile\ #1}%

\usepackage{mathtools}
\usepackage{enumitem}

\definecolor{bluegray}{rgb}{0.4, 0.6, 0.8}
\definecolor{brightturquoise}{rgb}{0.03, 0.91, 0.87}
\definecolor{cadetblue}{rgb}{0.37, 0.62, 0.63}
\definecolor{NVblue}{rgb}{0.07, 0.12, 0.83}

\definecolor{BUred}{rgb}{0.8, 0.0, 0.0}

\definecolor{caribbeangreen}{rgb}{0.0, 0.8, 0.6}
\definecolor{darkpastelgreen}{rgb}{0.01, 0.75, 0.24}
\definecolor{emerald}{rgb}{0.31, 0.78, 0.47}

\definecolor{ourlightgray}{rgb}{0.9, 0.9, 0.9}
\definecolor{ourdarkgray}{rgb}{0.77, 0.77, 0.77}

\usepackage{hyperref}
\hypersetup{pagebackref=true,breaklinks=true,colorlinks=true,citecolor=NVblue,linkcolor=BUred,bookmarks=false}

\usepackage[capitalize]{cleveref}
\crefname{section}{Sec.}{Secs.}
\Crefname{section}{Section}{Sections}
\Crefname{table}{Table}{Tables}
\crefname{table}{Tab.}{Tabs.}

\def\cvprPaperID{11648} 
\def\confName{CVPR}
\def\confYear{2023}

\begin{document}


\title{CoralStyleCLIP: Co-optimized Region and Layer Selection for Image Editing}

\author{Ambareesh Revanur$^{*}$\quad
Debraj Basu\quad Shradha Agrawal\quad Dhwanit Agarwal\quad Deepak Pai\\
{Adobe Inc.}\\
{\tt\small $^{*}$arevanur@adobe.com}\\
}

\twocolumn[{
\maketitle
\renewcommand\twocolumn[1][]{#1}
\begin{center}
  \includegraphics[width=1\linewidth]{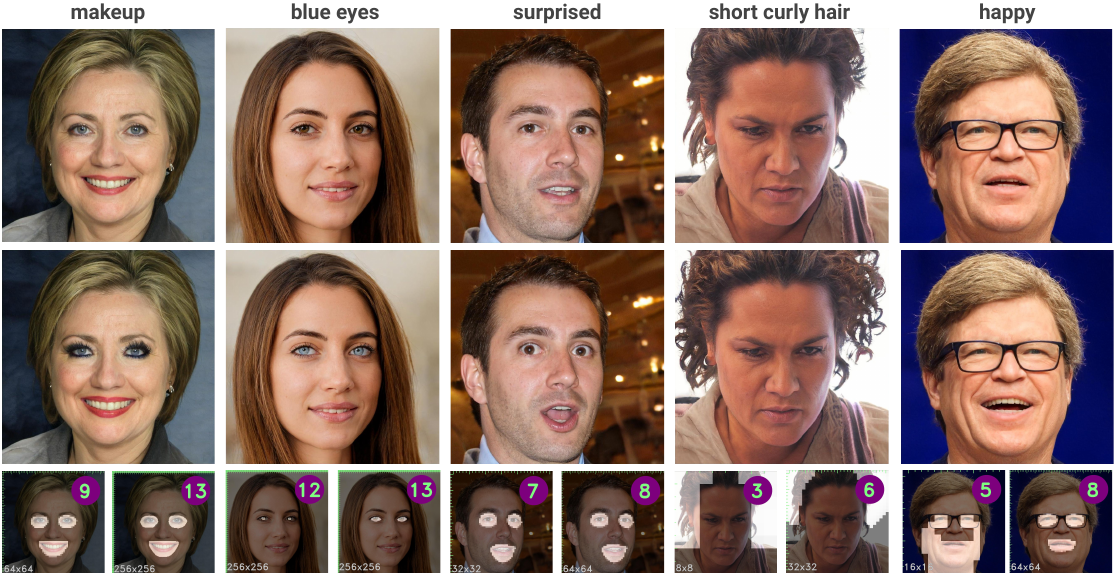}
  \captionsetup{type=figure,font=small}
  \vspace{-10pt}
  \caption{The original and edited images using CoralStyleCLIP are shown in the first and second rows of images, respectively. The bottom row shows the regions and StyleGAN2 layer numbers automatically selected for editing. The driving text prompts are above every column.
  }
  \label{fig:teaser}
\end{center}
}]

\begin{abstract}
\vspace{-0mm}

Edit fidelity is a significant issue in open-world controllable generative image editing. Recently, CLIP-based approaches have traded off simplicity to alleviate these problems by introducing spatial attention in a handpicked layer of a StyleGAN. In this paper, we propose CoralStyleCLIP, which incorporates a multi-layer attention-guided blending strategy in the feature space of StyleGAN2 for obtaining high-fidelity edits. We propose multiple forms of our co-optimized region and layer selection strategy to demonstrate the  variation of time complexity with the quality of edits over different architectural intricacies while preserving simplicity. We conduct extensive experimental analysis and benchmark our method against state-of-the-art CLIP-based methods. Our findings suggest that CoralStyleCLIP results in high-quality edits while preserving the ease of use. 

\end{abstract}

\vspace{-9mm}
\section{Introduction}
\vspace{-2mm}
\label{sec:intro}
\input{introduction}

\section{Related Work}
\label{sec:related_work}
\input{related_work}

\section{Approach}
\label{sec:approach}
\input{approach}

\section{Experiments}
\label{sec:expmt}
\input{experiments}

\section{Conclusion}
\label{sec:conclusion}
\input{conclusion}

\clearpage

\appendix
\section{Notation}
\label{app:notation}
\input{notation}

\section{Pseudocode for \Sectionref{approach}}
\label{app:pseudo}
\input{pseudocode}

\section{CLIP loss for semantic alignment}
\label{app:ablations}
\input{cliploss.tex}

\section{Architecture diagram}
\label{app:arch}
\input{architecture_diagram}

\section{Additional experiment details}
\label{app:addexpmt}
\input{additional_experiment_details}

\section{Additional results}
\label{app:addresults}
\input{additional_results}

\section{CORAL v/s multi-layer FEAT}
\label{app:multilayer_feat}
\input{multilayer_feat}

\section{Future applications}
\label{app:impact}
\input{impact}

\clearpage
{\small
\bibliographystyle{ieee_fullname}
\bibliography{bibliography_coralstyleclip}
}

\end{document}

%% file: headers.tex
\usepackage[T1]{fontenc}
\usepackage{graphicx}
\usepackage{array}
\usepackage{mdwmath}
\usepackage{mdwtab}
\usepackage{eqparbox}
\usepackage{fixltx2e}
\usepackage{url}
\usepackage{lipsum}
\usepackage{multirow}
\usepackage{subcaption}
\usepackage{amsfonts,amssymb,amsthm}
\usepackage{xspace}
\usepackage{xcolor}
\usepackage{bm}
\usepackage{bbm}
\usepackage{booktabs}
\usepackage{etoolbox}
\usepackage{makecell}
\usepackage{nicefrac}
\usepackage{textcomp}
\usepackage{stmaryrd}
\usepackage{mathrsfs}
\usepackage{paralist}
\usepackage{comment}

\usepackage[font=footnotesize,labelsep=space,labelfont=bf]{caption}

\usepackage{url}

\makeatletter

\newcommand{\Rmnum}[1]{\expandafter\@slowromancap\romannumeral #1@}
\makeatother

\newtheorem*{theorem*}{Theorem}
\newtheorem*{lemma*}{Lemma}
\newtheorem*{cor*}{Corollary}

\newcommand{\namedref}[2]{\hyperref[#2]{#1~\ref*{#2}}}

\newcommand{\Algorithmref}[1]{\namedref{Algorithm}{algo:#1}}

\newcommand{\Sectionref}[1]{\namedref{Section}{sec:#1}}
\newcommand{\Subsectionref}[1]{\namedref{Section}{subsec:#1}}

\newcommand{\Appendixref}[1]{\namedref{Appendix}{app:#1}}

\newcommand{\Tableref}[1]{\namedref{Table}{tab:#1}}
\newcommand{\Figureref}[1]{\namedref{Figure}{fig:#1}}

\newcommand{\Pageref}[1]{\hyperref[#1]{page~\pageref*{#1}}}

\definecolor{darkred}{rgb}{0.5, 0, 0} 
\definecolor{darkblue}{rgb}{0,0,0.5}

\newcommand{\bx}{\ensuremath{{\bf x}}\xspace}

\newcommand{\by}{\ensuremath{{\bf y}}\xspace}

\newcommand{\N}{\ensuremath{\mathcal{N}}\xspace}

\newcommand{\R}{\ensuremath{\mathbb{R}}\xspace}

\renewcommand{\paragraph}[1]{\smallskip\noindent{\bf #1}~}

%% file: introduction.tex
Controlling smooth semantic edits to photorealistic images~\cite{styleclip,age,interface,video} synthesized by well-known Generative Adversarial Networks (GANs)~\cite{gan,sg1,sg2} has become simplified with guidance from independently trained contrastive models such as CLIP~\cite{clip}. Using natural language as a rich medium of instruction for open-world image synthesis~\cite{Reed2016GenerativeAT,zhang2017stackgan,Zhang2019StackGANRI,Xu2018AttnGANFT} and editing~\cite{Dong2017SemanticIS, Nam2018TextAdaptiveGA, Liu2020DescribeWT,xia2020tedigan,nada,Lee2022SoundGuidedSI} has addressed many drawbacks of previously proposed methods.

\begin{figure}[h]
\begin{center}
\includegraphics[width=0.98\linewidth]{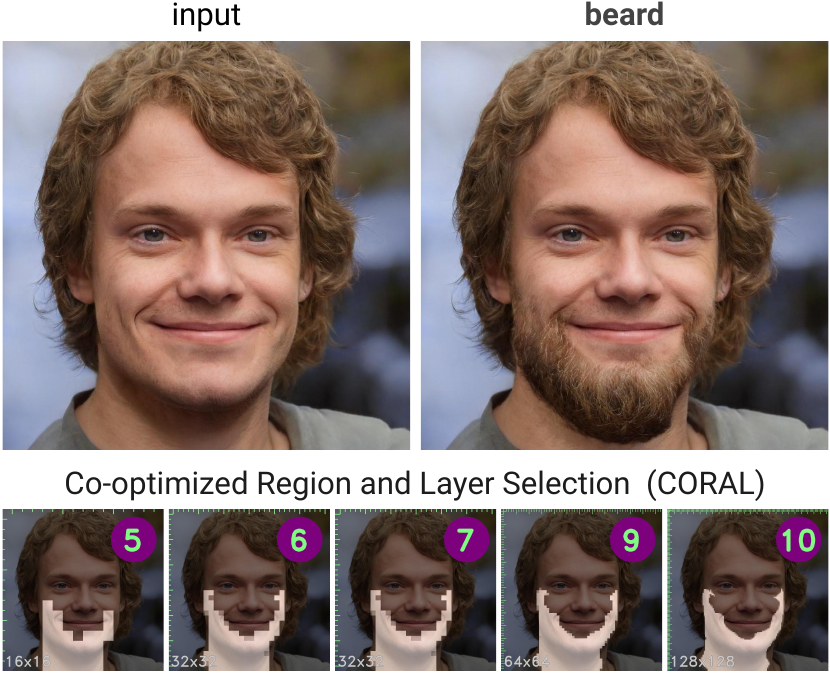}
\end{center}
\vspace{-0.5cm}
   \caption{For achieving {\it beard}, CORAL selects appropriate regions in layers 5-10 for carrying out the required coarse edits in early layers and finer texture edits in later layers}
\label{fig:img1}
\vspace{-0.5cm}
\end{figure}

As first demonstrated by StyleCLIP~\cite{styleclip}, the requirements for large amounts of annotated data~\cite{manigan} and manual efforts~\cite{ganspace,stylespace} were considerably alleviated. Furthermore, the range of possible edits that were achievable significantly improved~\cite{styleclip}. The underlying theme of related approaches involves CLIP-driven exploration~\cite{styleclip,stylemc,feat} of the intermediate disentangled {\it latent spaces} of the GANs. 

It is well understood by now that manipulating the latent code of a StyleGAN for aligning with a {\it text prompt} can be computationally intense, as seen in StyleCLIP latent optimization~\cite{styleclip}, as well as the latent mapper methods~\cite{styleclip}. This presents a trade-off between the complexity and quality of edits leveraged by StyleCLIP global directions~\cite{styleclip} and StyleMC~\cite{stylemc}. 

In addition, these methods often result in undesirable edits to unexpected regions of an image (see \cite{feat}), addressed to some extent by FEAT~\cite{feat}. However, FEAT requires manual intervention, as described in \Sectionref{related_work}, and involves significant training complexity of the order of 
hours\footnote{
With no official implementation available, we present comparisons with our reimplementation of FEAT denoted by FEAT\textsuperscript{*} in this paper.
}.

\noindent {\bf Contributions.} In this paper, we propose CoralStyleCLIP, which addresses these challenges by combining the ease of use~\cite{styleclip} with efficient~\cite{stylemc} high fidelity edits~\cite{feat} into our approach. 
In particular, we propose a novel strategy which, for a given text prompt, jointly learns both the appropriate direction of traversal in the latent space, as well as which spatial regions to edit in every layer of the StyleGAN2~\cite{sg2} (see \Figureref{teaser}, \Figureref{img1}) without any mediation. 

Our approach overcomes the need for manual effort in selecting an appropriate layer for FEAT by incorporating multi-layer feature blending to enable the joint learning process. As a result, the edits are very accurate, rendering our method simple and effective. 

The co-optimized regions and layers jointly learned with appropriate latent edits typically select earlier layers for enacting coarse edits, such as shape and structural, compared to finer edits, such as color and texture, which are usually orchestrated through the latter layers of the StyleGAN2. 

To alleviate the time complexity, we implement this strategy for {\it segment selection} (see \Subsectionref{coral}), where we jointly learn a {\it global direction}~\cite{styleclip,stylemc} in the $\mathcal{W}^+$ space and limit the predicted areas of interest at every layer to segments from a pre-trained segmentation network. Doing so reduces the learning complexity significantly (see \Tableref{quant_comparison}), albeit with potential pitfalls discussed in \Subsectionref{results}. We mitigate these pitfalls with a jointly trained \textit{attention network} where we relax the areas of interest at every layer to spatial masks predicted by the network (see \Subsectionref{coral}). As a result, the training time increases from a few minutes to about an hour while improving the quality of edits compared to the {\it segment selection} approach.

\vspace{1mm}
\noindent In summary, our contributions are as follows:

\vspace{-1mm}
\begin{itemize}
    
    \item We propose a novel multi-layer blending strategy that attends to features selectively at the appropriate StyleGAN layer with minimal hand-holding.
    \vspace{-1mm}
    \item  A CORAL variant based on {\it segment selection} demonstrates high edit quality at a fraction of time cost.
    \vspace{-1mm}
    \item Through extensive empirical analysis, we find that CORAL outperforms
    recent state-of-the-art methods and is better equipped to handle complex text prompts.
    
\end{itemize}

%% file: related_work.tex
The use of generative models for high-quality image synthesis and manipulation has a rich history~\cite{stylegan_survey,biggan,progan}. In particular, the disentangled latent spaces of StyleGAN provide robust interpretable controls for editing valuable semantic attributes of an image~\cite{stylespace,ganspace,collins2020editing, interface,tewari2020stylerig,anima2020disentangled,zhou2021fact,clip2stylegan}. Desirable changes to attributes of interest were previously brought out by discovering the relevant channels~\cite{stylespace} and curating principal components~\cite{ganspace} either through manual inspection or otherwise driven by data-hungry attribute predictors. 

StyleFlow~\cite{styleflow} leverages normalizing flows to perform conditional exploration of a pre-trained StyleGAN for attribute-conditioned image sampling and editing. By learning to encode the rich local semantics of images into multi-dimensional latent spaces with spatial dimension, StyleMapGAN~\cite{stylemapgan} demonstrates improved inversion quality and the benefits of spatially aware latent code interpolation between source and target images for editing purposes.  
The advent of CLIP~\cite{clip} has re-ignited interest in open domain attribute conditioned synthesis of images~\cite{styleclip,clip2latent,tigan}. Text-driven edits have considerably reduced both the time and effort required for editing images and extended the range of possible edits significantly~\cite{styleclip}, all the more with increased interest in diffusion models~\cite{dalle, ramesh_hierarchy,glide}. 

The disentangled nature of the latent spaces of StyleGAN has facilitated heuristics such as a fixed global direction in StyleCLIP~\cite{styleclip} and, more recently, StyleMC~\cite{stylemc}. For training efficiency, StyleMC performs CLIP-driven optimization on the image generated at a low-resolution layer of the StyleGAN. Unfortunately, this limits the range of edits to only those possible by manipulating latent codes at the earlier layers.

For ameliorating edits in unexpected regions of an image, strategies for blending latent features have been an emerging theme in many recent papers~\cite{stylefusion,stylemapgan,feat,blenddiff}. \cite{stylemapgan,feat,blenddiff} interpolate spatial features more explicitly. In contrast, StyleFusion~\cite{stylefusion} realizes similar objectives through blended latent code extracted using a fusion network that combines disjoint semantic attributes from multiple images into a single photorealistic image.

\begin{figure}
\begin{center}
\includegraphics[width=\linewidth]{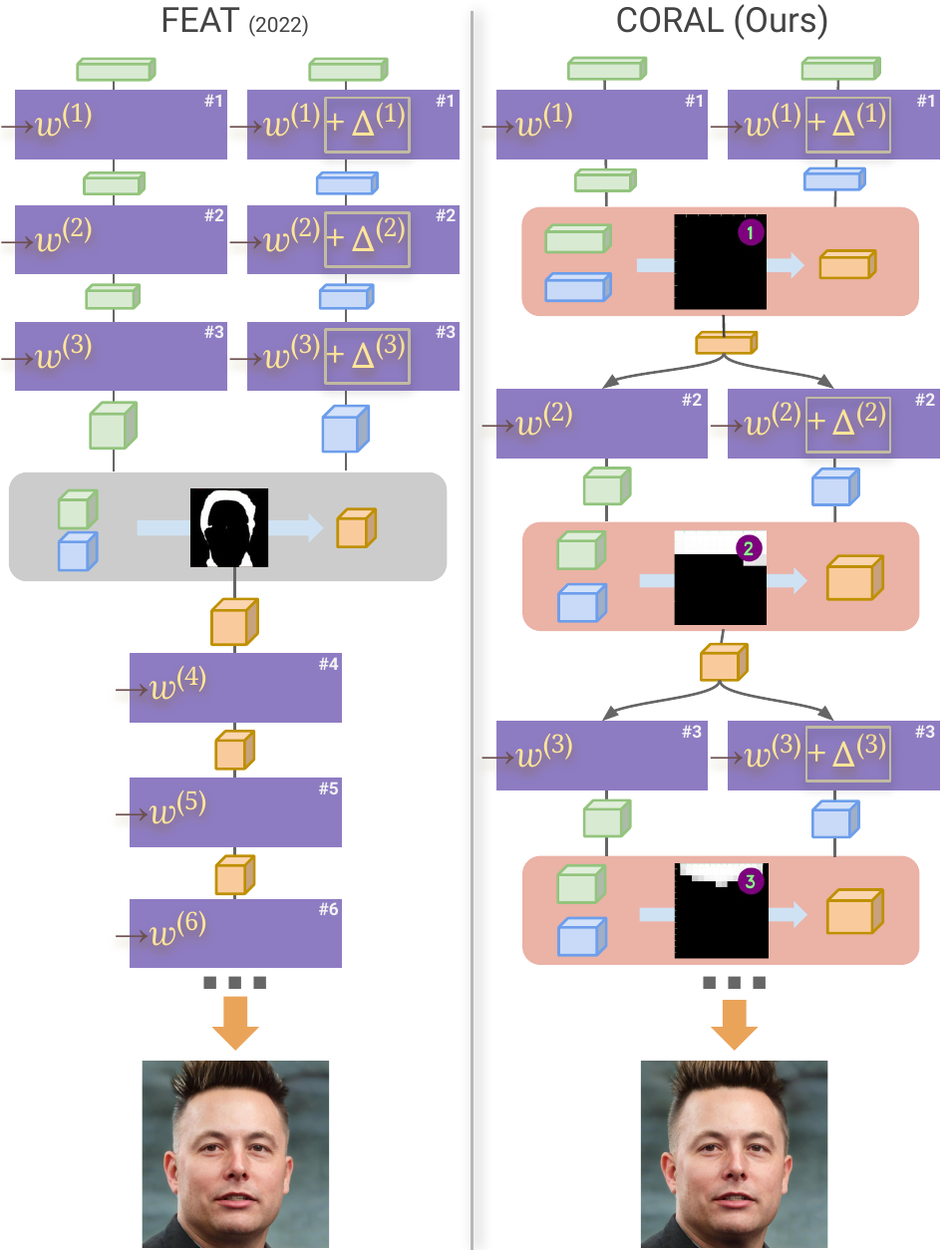}
\end{center}
\vspace{-0.5cm}
   \caption{Comparison of FEAT~\cite{feat} with CORAL. In FEAT (left), the spatial features are blended at a carefully hand-picked layer $l$.
   CORAL (right) performs multi-layer blending with custom edit regions per layer. 
   }
\label{fig:img2}
\vspace{-0.5cm}
\end{figure}

Our work is most closely related to \cite{feat,stylemc, styleclip}. FEAT~\cite{feat} reduces undesirable edits by imposing sparsity in the number of spatial features modified by StyleCLIP at a manually selected layer $l$ of the StyleGAN2. FEAT edits layers $\leq l$ using a non-linear latent mapper, while the attention network emits a spatial mask for interpolating edited spatial features at layer $l$ with original spatial features at the same layer (see \Figureref{img2}). At the cost of training time and convenience, FEAT achieves high-fidelity edits. If the blending layer is not carefully selected, the edits can be significantly poor, as shown in \cite{feat} and \Figureref{67}. Furthermore, FEAT enacts inferior edits when presented with multi-faceted prompts (see \Figureref{complexprompt}). In Suppl., we also discuss how CORAL is different from a multi-layer extension of FEAT.

Furthermore, we argue that the required edits for aligning with a given text prompt arise from multiple layers of the StyleGAN2, necessitating a multi-layer feature interpolation mechanism (see \Subsectionref{blending}). Our method percolates meaningful edits from the current layer onto subsequent layers, with restrictions on the number of spatial edits customized for each layer. As a result, we can automatically select the correct layers and regions for editing an image.

To correctly identify the region of interest at every layer, we discuss a lightweight segment-selection scheme (see \Subsectionref{coral}) and contrast this with an involved convolutional mask prediction model motivated by FEAT. Recently, SAM~\cite{SAM} accomplished superior GAN inversion at the cost of editability by leveraging different latent spaces of the StyleGAN2 in a spatially adaptive manner. However, the edits performed on the inverted latent codes continue to modify irrelevant image regions and could benefit from CORAL (see \Sectionref{approach}). 

With a focus on convenience and fidelity, CoralStyleCLIP learns global directions at every layer of the StyleGAN2, as done in \cite{stylemc}, and exhibits high-quality edits with a significant reduction in the training time and manual effort (see \Tableref{quant_comparison}). Borrowing inspiration from \cite{styleclip}, we also implement our co-optimized region and layer selection strategies for a non-linear mapper-based latent edit and demonstrate additional customized and high fidelity edits. 

%% file: approach.tex
An image edit is often spatially localized to a specific region of interest.  
For example, edits corresponding to the {\it mohawk} text prompt should affect only the hair region of the portrait image while preventing edits in other parts.
In this work, we learn a latent edit vector and a soft binary mask at every layer of a StyleGAN2 to accurately edit the image according to the input text prompt. We achieve this by training them end-to-end while respecting the challenging but desirable minimal overall edit area constraint. Following a brief revisit to the StyleGAN architecture, we introduce two simple yet effective strategies to determine the region of interest given a text prompt. Finally, we introduce a novel multi-layer blending strategy that is vital for achieving high fidelity minimal edits.

\subsection{Background}
\label{subsec:background}

StyleGAN2~\cite{sg2} is a state-of-the-art model trained for generating high-resolution images typically of sizes $1024\times1024$ or $512\times512$.
The network consists of a mapper module that maps a random vector $z \in \mathcal{Z} \sim {\mathcal{N}(0, 1)}$ to a vector in $w\in \mathcal{W}$ space via a multi-layer perceptron (MLP), and a generator module comprising 18 convolutional blocks. 

The $\mathcal{W}^+$ space, first defined by \cite{im2style}, is a concatenation of 18 different $w^{(l)}$ vectors where $l\in\{1,2,\ldots,18\}$. The $w^{(l)}$ instance in $\mathcal{W}^+$-space is first transformed through a layer-specific affine operation to obtain {\it stylecode} $s^{(l)} \in \mathcal{S}$, at all layers of the generator module. The input to the generator module is a learned tensor of $4\times 4$ resolution. It is gradually increased to a resolution of $1024\times1024$ as the input tensor is passed down through the layers of the generator.

We denote the constant input tensor as $c$ and the feature obtained at a layer $l$ as $f^{(l)}$. Further, we denote the $\mathcal{W}^+$ code at layer $l$ as the $w^{(l)}$ and a layer in generator module as $\Phi^{(l)}$. Therefore, $f^{(l)}$ can be expressed as $f^{(l)} = \Phi^{(l)} (f^{(l-1)}, w^{(l)})$, where $l\in \{1, 2, \cdots, 18\},\,c = f^{(0)}$ and the generated image $I=\sum_{l=1}^{18}RGB^{(l)}(f^{(l)})$.

In our work, we aim to find a latent vector $\Delta^{(l)}$ in the $\mathcal{W}^+$ such that the image generated by the latent code $w^{(l)}+\Delta^{(l)}$ applied to every layer of generator results in an edited image $I^*$. For simplicity, we denote $f^*$ and $w^*=w+\Delta$ as edited features and $\mathcal{W}^+$ latent code, respectively. Therefore, we have $f^{*(l)} = \Phi^{(l)} (f^{*(l-1)}, w^{(l)}+\Delta^{(l)})$. A recent study showed that StyleGAN2 learns global attributes such as position in earlier layers, structural changes in middle layers, and appearance changes (e.g., color) in the final set of layers \cite{sg1,xia2020tedigan}. However, determining the right set of layers for a given text prompt is challenging and has been explored only empirically in FEAT \cite{feat}.

\subsection{Co-optimized region and layer selection (CORAL)}
\label{subsec:coral}
We aim to edit the image to match the text prompt with minimal changes.
To this end, the first step is correctly identifying the region of interest. Further, given the diversity and richness of latent space at each layer in the generator, we posit that the edits to the image can come from multiple layers of the StyleGAN2 generator. 

To address both requirements, we introduce CORAL, a co-optimized region and layer selection mechanism. In CORAL, we propose two simple-yet-effective approaches for learning a soft binary mask $m^{(l)} \in [0,1]^{f^{(l)}_{dim}}$ at every layer of the generator module with the same height and width dimensions as the feature resolution at the given layer.

\begin{figure}[h!]
\begin{center}
\includegraphics[width=\linewidth]{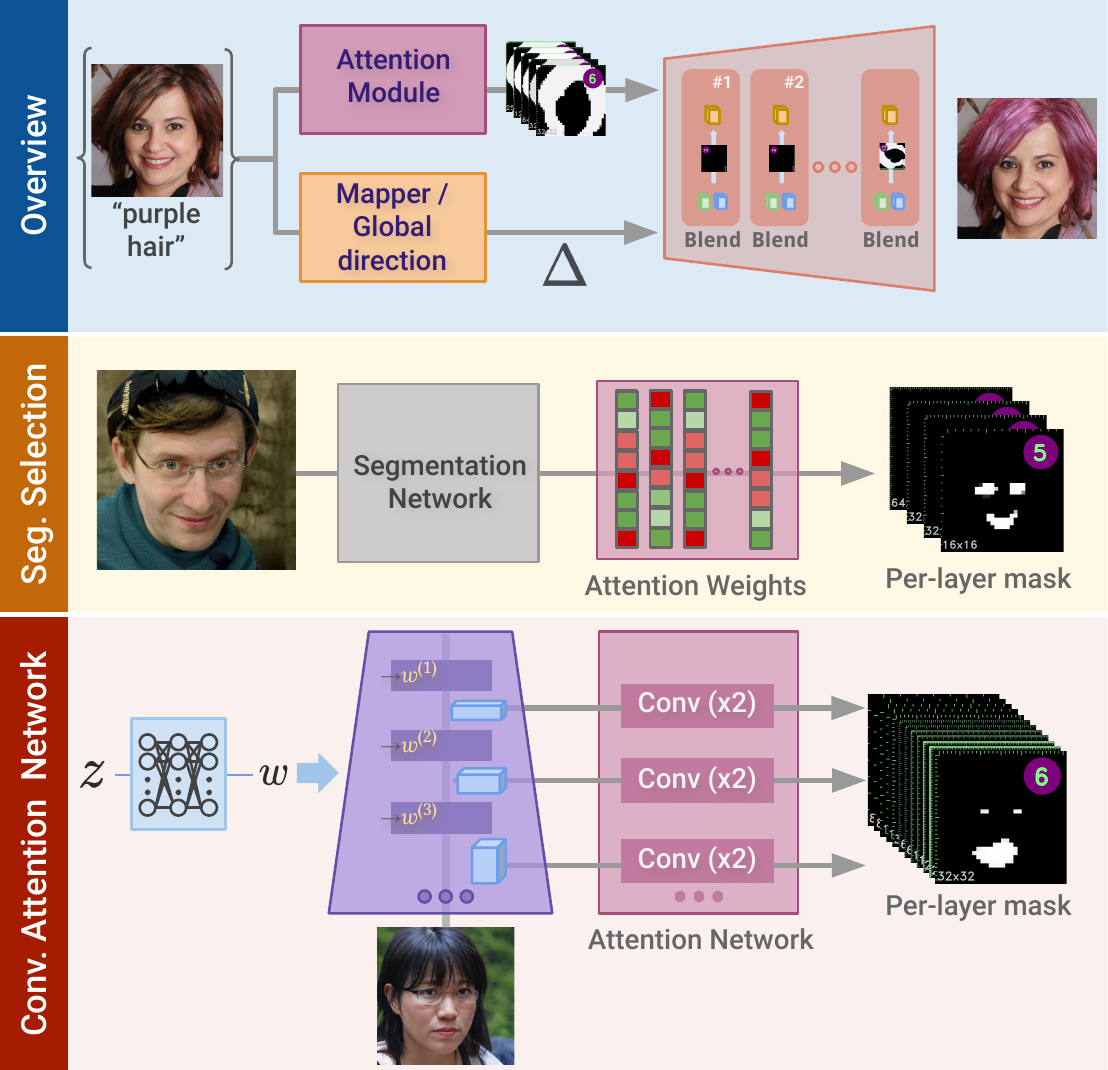}
\end{center}
\vspace{-0.5cm}
   \caption{ Overview of CoralStyleCLIP. The only trainable components are the attention module and the mapper/global direction. Two different variants of the attention module are summarized as segment selection and convolutional attention network (see \Subsectionref{coral} for more details). 
   }
\label{fig:approach}
\vspace{-0.5cm}
\end{figure}

\vspace{1mm}

\noindent \textbf{CORAL based on segment-selection.} 
We can use any off-the-shelf pre-trained semantic segmentation network to determine the region of interest in this approach. Intuitively, existing image segmentation networks generally capture semantic parts of the image that we are interested in editing, such as eyes, mouth, and lips. Therefore in many cases, this problem can be posed as selecting the appropriate segments. To achieve this, we introduce a matrix $e$ of dimension $P \times 18$ where $P$ is the number of classes predicted by the segmentation network. Each entry in the matrix $e$ is in the range $[0, 1]$, where 1 represents a confident segment selection for the given text prompt $t$.

The matrix $e$ is converted into a spatial mask $m^{(l)}$ by masking the segments with the confidence values and resizing the segmentation map to the resolution of the feature maps at each layer. In the training phase, the parameters in the matrix $e$ are trained after applying a sigmoid, and during inference, we apply a prompt-specific threshold $\tau_t$ to the sigmoid.
As depicted in \Figureref{approach}, the only trainable parameters in this pipeline are $e$. Therefore, this can achieve desirable edits with high accuracy up to 8x faster than FEAT~\cite{feat}.

\vspace{1mm}

\noindent \textbf{CORAL based on convolutional attention network.} Segment-selection-based CORAL is limited by the segments available in the pre-trained network. As shown in results \Figureref{67}-F, the segment-selection method is prone to over-selection or under-selection of the region of interest. To overcome this limitation, we implement an attention network that directly predicts the masks $m^{(l)}$ at every layer of the generator as shown in \Figureref{approach}. In this architecture, we obtain a mask with the exact resolution as that of the corresponding feature in the layer. Unlike FEAT, we hypothesize that the mask at a layer $l$ should depend only on the features $f^{(l)}$ available at the current layer since we are interested in predicting the mask at every layer. 

Despite incurring higher training costs from having to learn the convolutional layers, the masks produced with this approach are smoother and avoid over/under-selection issues by accurately predicting the correct region of interest. 

\subsection{Multi-layer feedforwarded feature blending}
\label{subsec:blending}

CORAL produces soft binary masks $m^{(l)}$ at every layer of the generator module. These masks blend features such that the features corresponding to the confident regions are borrowed from features $f^*$ generated with updated style code, and on similar lines, features from non-confident regions are borrowed from original features $f$ of the unedited image. This ensures that we only modify the regions corresponding to the text prompt and prevents modifications of non-masked regions. Unfortunately, a 0-mask (completely black mask) at any layer would throw away any updated feature information from the previous layers and would propagate the original features $f$ from that point onward.

To prevent this bottleneck, we design a novel multi-layer feature blending strategy (see \Figureref{img2}) that utilizes a parallel pathway where the feature obtained from layer $l-1$ is passed through the generated block $\Phi$ twice - once with the original latent code $w$ and another pathway with updated latent code $w+\Delta$ to obtain two feature sets for blending. The former feature can be viewed as a feature that is not edited but has all the information propagated from previous layers. 
The multi-layer blending strategy expressed in \eqref{mlbs}, ensures that no feature information is lost along the way.\vspace{-0.1cm}
\begin{align}
    \widehat{f^{*(l)}} &=  \Phi^{(l)}(f^{*(l-1)}, w^{(l)}+\Delta^{(l)})\nonumber\\
    \widehat{f^{(l)}} &=  \Phi^{(l)}(f^{*(l-1)}, w^{(l)})\nonumber\\
    f^{*(l)} &=m^{(l)} \odot \widehat{f^{*(l)}} + \newline (1-m^{(l)}) \odot  \widehat{f^{(l)}}\label{mlbs}
\end{align}

Intuitively, when the mask is completely blank (which is often desirable to keep the edits to a minimum), the features are feedforwarded simply with edits from previous layers. 

\subsection{Types of latent edits}
\label{subsec:lated}
For a given convolutional layer $l$, when the learned latent edit $\|\Delta^{(l)}\|>0$, the corresponding feature $\widehat{f^{*(l)}}$ in \eqref{mlbs} incorporates attributes which are desirable for semantic alignment with the given text prompt. The mask $m^{(l)}$ counteracts possible undesirable artifacts through a region-of-interest-aware interpolation strategy. 

The $\Delta^{(l)}$ by itself is, however, well studied in \cite{styleclip,stylemc}, both of which identify a single global direction that can semantically edit images for a given text prompt. Such a simple parameterization does result in accurate edits for simple text prompts, as discussed in \cite{styleclip}. 

Our findings suggest that training time is significantly reduced for prompts where a global direction can affect desirable changes. However, a more involved image-dependent non-linear mapper model $g(\cdot)$ as a function of $w^{(l)}$ at every layer can affect such changes with higher precision. 

Therefore, we implemented CORAL for both versions of latent edits: {\bf (i)} {\it global direction}; {\bf (ii)} {\it latent mapper}. The latent mapper $g(\cdot)$ is an MLP-based model along the lines of \cite[Section 5]{styleclip}, where the $w^{(l)}$ are split into three groups: coarse ($l$ in 1 to 4), medium ($l$ in 5 to 8) and fine ($l$ in 9 to 18); and each of these groups is processed by a different MLP\footnote{Unlike in \cite{styleclip}, we remove the \texttt{LeakyReLU} activation after the final fully connected layer, as it empirically expedites the optimization.}. 
Our multi-layer feature blending mechanism is independent of the parametrization of the latent edit, which is jointly learned with the mask $m^{(l)}$ predictors. 

\subsection{Loss formulation}We now describe our proposed methods' training strategy and loss formulation. We are given a text prompt $t$ and an image with corresponding $\mathcal{W}^+$ code $w$. The goal is to find the right region of interest using a CORAL framework and determine the latent vector to help with the image edit. The only trainable components in our approach are the latent vector $\Delta$ and the parameters in the CORAL framework. In the case of segment selection, the only trainable component in CORAL is the matrix, and in the case of convolutional attention networks, the Conv layers in the attention network are trainable.

\noindent \textbf{CLIP loss}: The first key loss component is the CLIP loss originally proposed in StyleCLIP~\cite{styleclip}. The idea is to use the pre-trained CLIP model to edit the latent vector such that the embedding of the image $I^*$ produced aligns with the embedding of the text prompt $t$ in CLIP’s latent space. 

In addition we also synthesize the image $\widetilde{I}$, by setting $m_{i,j}^{(l)}=1\,\,\forall{i,j,l}$ in \eqref{mlbs} and compute its CLIP loss. To understand this, we can envision $I^*$ as a sophisticated non-linear interpolation between $I_0$ and $\widetilde{I}$ using strategies given in \eqref{mlbs}. Here $I_0$ is the original unedited image.

By simply imposing a CLIP loss on $I^*$, $\widetilde{I}$ remains unrestricted and can potentially contain undesirable artifacts, as long as $I^*$ aligns with the text prompt $t$. However, our region selectors in \Subsectionref{coral} derive their supervision from $\widetilde{I}$ and might also learn to include these artifacts. Our final semantic alignment loss is as follows:
\begin{small}
\begin{equation}
\label{eq:clip}
\mathcal{L}_{clip} = \frac{1}{2}\left(D_{\text{CLIP}}(I^*, t) + D_{\text{CLIP}}(\widetilde{I}, t)\right)
\end{equation}
\end{small}

\noindent {\bf $\text{L}_2$ loss}: Controlled perturbations to the latent spaces of a StyleGAN2 can result in smooth semantic changes to the generated image. As a result, we optimize the squared Euclidean norm of $\Delta$, i.e., $\mathcal{L}_{l_2}=\|\Delta\|_2^2$, in the $\mathcal{W}^+$ space to prefer latent edits with smaller $l_2$ norms. 

\noindent \textbf{ID loss}: In order to prevent changes to the identity of a person during image manipulation, we impose an ID loss $\mathcal{L}_{id} = 1 -  \langle\mathcal{R}(I^*), \mathcal{R}(I) \rangle$ using 
cosine similarity between the embeddings in the latent space of a pre-trained ArcFace network $\mathcal{R}$~\cite{arcface,encinstyle,styleclip,stylemc}.

\vspace{1mm}

\begin{figure*}[ht!]
\begin{center}
\includegraphics[width=\textwidth]{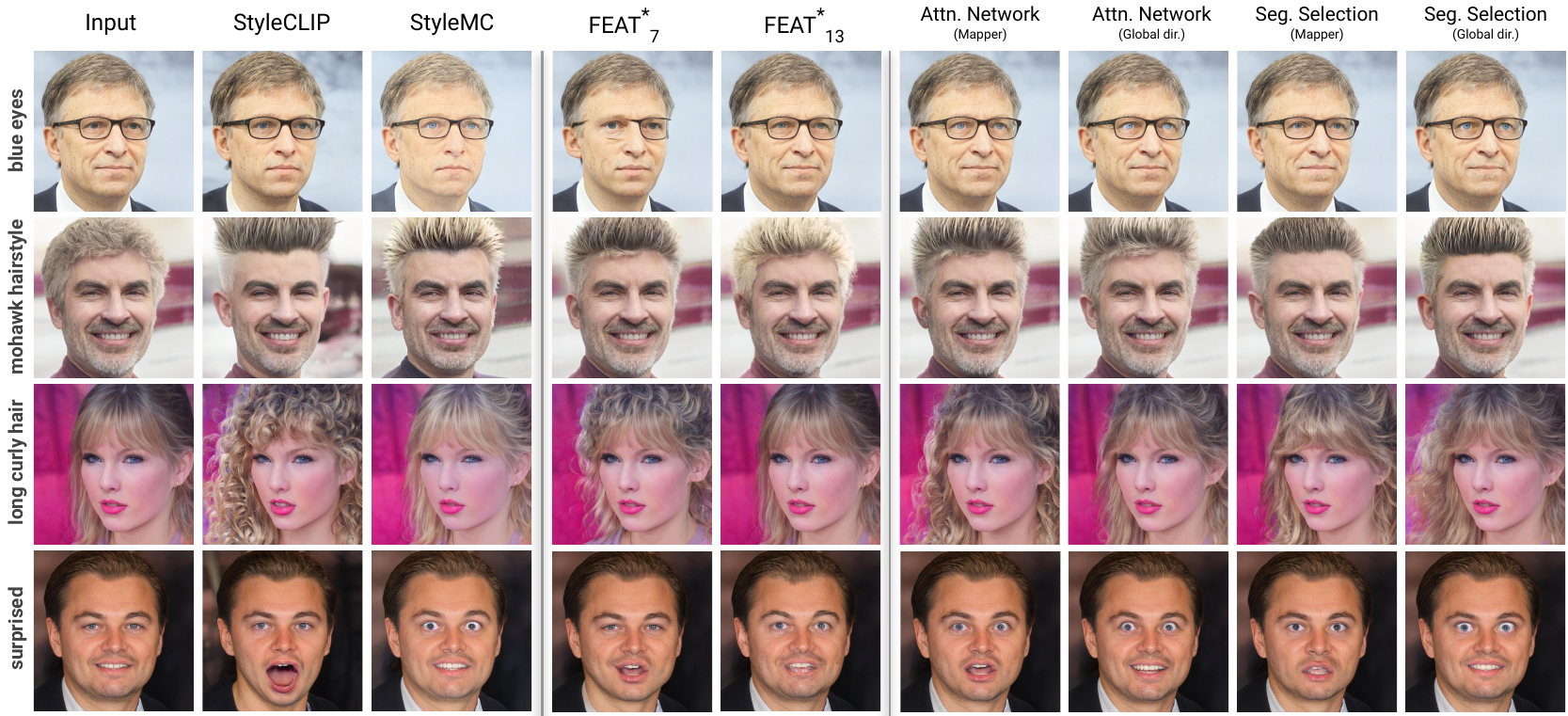}
\end{center}
\vspace{-0.5cm}
   \caption{Comparison of variants of CORAL differing in complexity with closely related FEAT\textsuperscript{*}~\cite{feat}, StyleCLIP mapper method~\cite{styleclip} and StyleMC~\cite{stylemc}}
\label{fig:compare}
\vspace{-0.5cm}
\end{figure*}

\noindent \textbf{Minimal edit-area constraint}: We encourage the network to find an edit with changes to compact image areas. In the case of segment selection, this is achieved by penalizing the CORAL matrix $e$ as follows:\vspace{-0.2cm}
\begin{small}
\begin{equation}
\label{eq:wreg_1}
\mathcal{L}^{ss}_{area} = \sum_{i,j} e_{i, j}
\end{equation}
\end{small}
In the case of a convolutional attention network, this is achieved by imposing the minimal edit constraint directly on the masks $m$ as follows:\vspace{-0.2cm}
\begin{small}
\begin{equation}
\label{eq:wreg_2}
\mathcal{L}^{can}_{area} = \sum_{l} n_l \Big(\sum_{i,j} m^{(l)}_{i,j}\Big)
\end{equation}
\end{small}
where $n_l$ is a normalizing constant defined per layer to account for the growing feature dimensions as the feature passes through the StyleGAN2 generator module. \vspace{1mm}

\noindent \textbf{Smoothness loss}: In the case of the convolutional attention network, it would be desirable to predict a smooth mask. This is achieved by imposing a total variation loss~\cite{feat}.
\vspace{-0.07cm}
\begin{small}
\begin{equation}
	\mathcal{L}_{tv} = \sum_{i,j,l} \Vert m^{(l)}_{i, j} - m^{(l)}_{i+1, j} \Vert_2^2 + \sum_{i,j,l} \Vert m^{(l)}_{i, j} - m^{(l)}_{i, j+1} \Vert_2^2
\end{equation}
\end{small}
In summary, the loss formulations for the segment selection and convolutional attention mechanisms are as follows:\vspace{-0.15cm}
\begin{small}
\begin{align}
    \label{eq:loss_segment_selection}
	\mathcal{L}_{ss} &= \mathcal{L}_{clip} + \lambda_{l_2}\mathcal{L}_{l_2} + \lambda_{id} \mathcal{L}_{id} + \lambda_{area} \mathcal{L}^{ss}_{area}\\
 \label{eq:loss_attention_network}
 \mathcal{L}_{can} &= \mathcal{L}_{clip} + \lambda_{l_2}\mathcal{L}_{l_2} + \lambda_{id} \mathcal{L}_{id} + \lambda_{area} \mathcal{L}^{can}_{area} + \lambda_{tv} \mathcal{L}_{tv}
\end{align}    
\end{small}

Both the CORAL module and the latent editor are optimized in an end-to-end fashion using the above losses.

%% file: experiments.tex
\begin{figure*}[ht!]
    \centering
    \includegraphics[width=\linewidth]{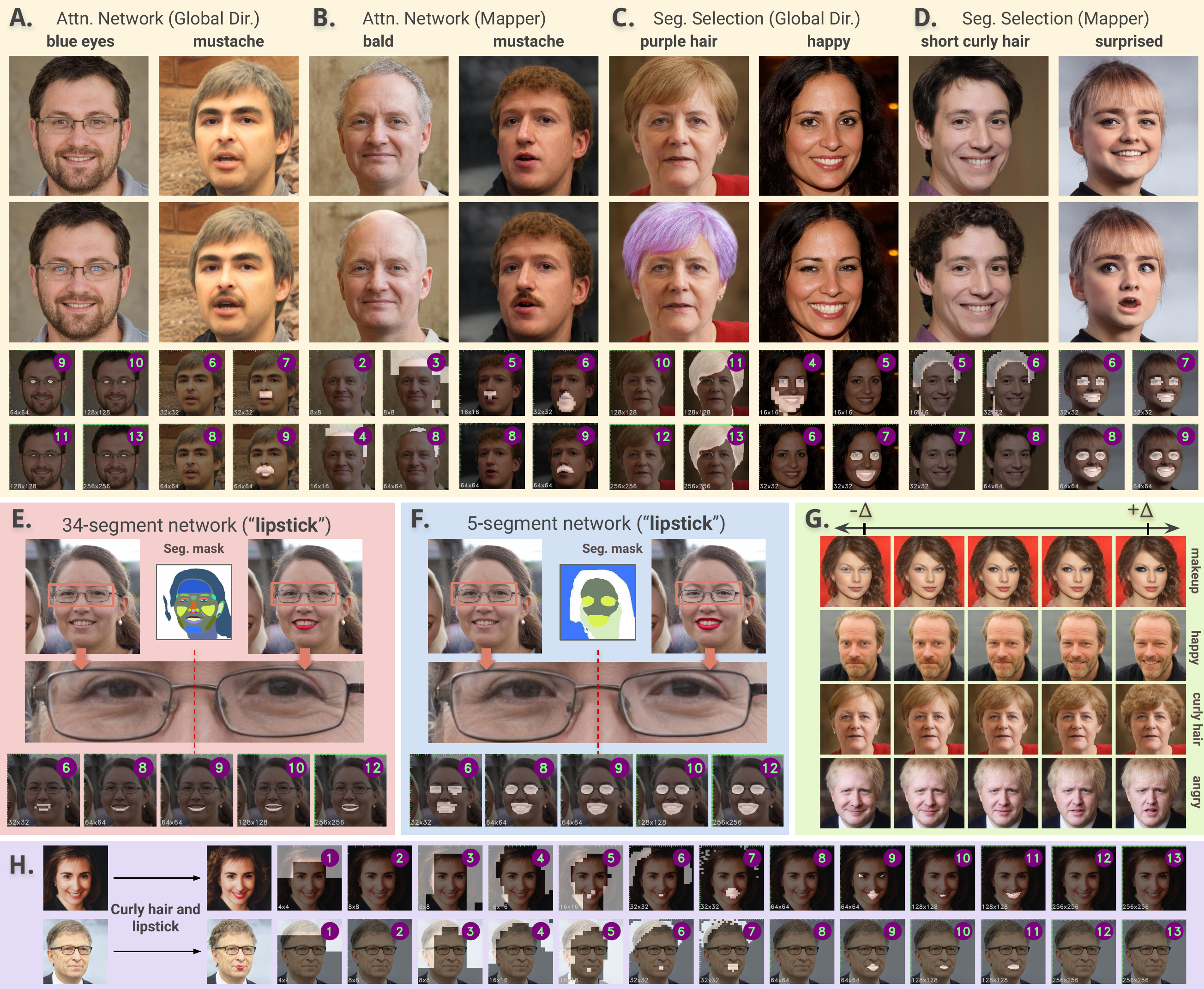}
    \vspace{-0.5cm}
    \caption{Each column in figures A to D demonstrates a text-driven edit on an input image along with the corresponding layers and regions selected. As a limitation of segment selection, we observe over-selection of the region of edit in figure F, which is absent in E. Figure G compares edits along both the positive and negative direction where we observe intuitive differences between removal and application of {\it makeup}, {\it happy} vs. {\it unhappy} and {\it curly} vs. {\it smooth hair}. Finally, Figure H demonstrates the edit regions selected by CORAL across different layers of the StyleGAN2 for a complex prompt. }
    \label{fig:67}
    \vspace{-0.5cm}
\end{figure*}

We evaluate CORAL mainly in the context of human faces and demonstrate high-quality edits to photo-realistic faces of size $1024\times 1024$ generated by a StyleGAN2 pre-trained on the FFHQ dataset~\cite{sg1}. We present additional results on \textit{sketch} and \textit{pixar} domains as well as Cars dataset \cite{stanfordcars} in Suppl. For both variants of CORAL in \Subsectionref{coral}, we compare edits from the {\it global direction} and {\it latent mapper} in  \Subsectionref{lated}. All hyperparameter configurations for \eqref{eq:loss_segment_selection} and \eqref{eq:loss_attention_network} are provided in Suppl.

\subsection{Training and inference}
\label{subsec:training}

\begin{figure}[t]
\begin{center}
\includegraphics[width=.95\columnwidth]{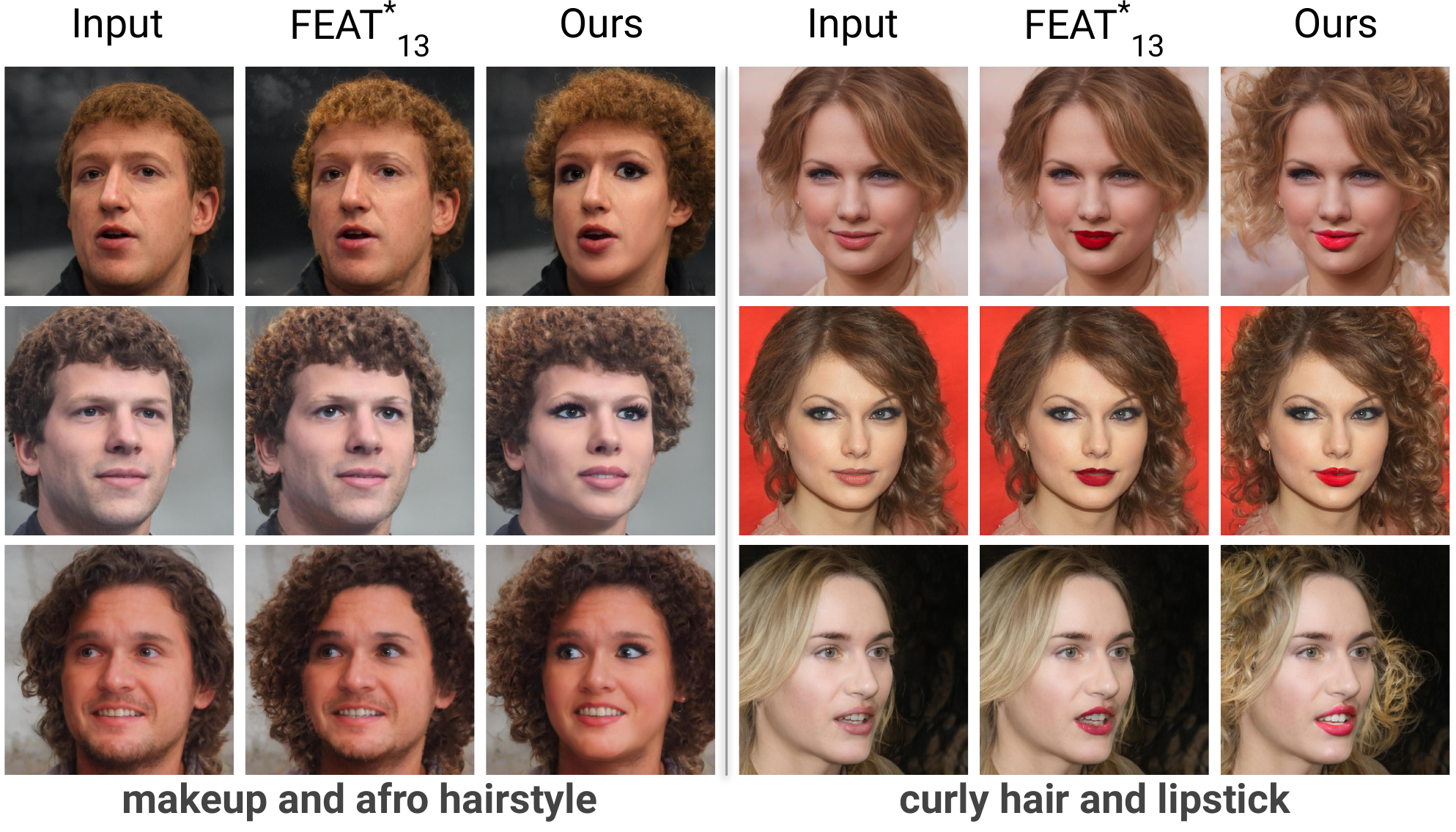}
\end{center}
\vspace{-0.6cm}
   \caption{Comparison of CORAL with $\text{FEAT}^*_{13}$ for multi-faceted prompts.
   }
\label{fig:complexprompt}
\vspace{-0.7cm}
\end{figure}

\vspace{-2mm}

The loss functions corresponding to the two different variants of CORAL are given by \eqref{eq:loss_segment_selection} and \eqref{eq:loss_attention_network}. Our experiments were conducted on one NVIDIA Tesla P40 24 GB GPU with a batch size of 3. The latent editor and CORAL modules are jointly optimized using an Adam optimizer~\cite{adam} while keeping the StyleGAN2 fixed. 

For a given text prompt $t$, a data point is given by a randomly sampled standard normal vector $z\sim \mathcal{Z}$ space, and the maximum number of iterations is set to 20,000. However, in \Tableref{quant_comparison}, we note that the training time required for achieving the desired quality of edit increases as we switch from segment-selection to a convolutional attention network, the same as in going from global direction edits to training a latent mapper. 

Furthermore, during inference, we limit the automatically selected regions for editing by setting $m^{(l)}\leftarrow m^{(l)} \odot \bm{1}\{m^{(l)} \geq \tau\} $ where $\tau$ is typically 0.85. For applying desirable edits and reversing them (see \Figureref{67}-G), we have a multiplying factor $\alpha\in[-1.5,1.5]$ for the edit direction $\Delta$. 

Out of the 18 convolutional blocks and the corresponding $w$ code per layer, our CORAL strategy and the latent edits, as well as edits from our baselines, are only performed on the first 13 layers, which are known to span coarse and fine controls over diverse attributes~\cite{stylespace} such as expressions, age, style and color of facial hair, and eyes, among others. 

\noindent{\bf Segment selection:}
Based on ideas from \cite{unsupseg}, a pre-trained mixture model is used for performing unsupervised semantic segmentation of the StyleGAN2 generated images into 5 classes per pixel. This model is then used to determine the region of interest with CORAL based on segment selection.
In \Figureref{67}-E, we also compare with CORAL for a weakly supervised 34 class DatasetGAN~\cite{datasetgan} network, trained on the features of StyleGAN2 network using few shot labels.

\noindent {\bf Attention network:}
At each convolutional block $l\in [1,2,\ldots,13]$ of the StyleGAN2, the attention network first applies 32 different $1\times 1$ convolutional filters upon the spatial features $f^{(l)}$ to reduce the number of channels to 32 followed by ReLU~\cite{relu} activation, after which another $1\times 1$ convolutional layer and sigmoid activation are applied to obtain $m^{(l)}$. We set $n_l=\nicefrac{1}{\text{size}[l]}$ in \eqref{eq:wreg_2}, where $\text{size}[l]$ is given based on the height and width of $f^{(l)}$, for example if the resolution is $32\times32$, then $n_l=\nicefrac{1}{32}$.

\subsection{Evaluation}
\label{subsec:eval}
Our method is most closely related to StyleCLIP~\cite{styleclip}, StyleMC~\cite{stylemc} and  FEAT~\cite{feat}. For a comparison with CORAL, we run the official implementation of the latent mapper technique of StyleCLIP, as well as a re-implementation of StyleMC\footnote{as per Section 3.2 of their paper} 
which optimizes a single global direction across multiple images, only for layers of the StyleGAN2 until resolution $256\times256$ (see \cite{stylemc}). 

Without an official implementation of FEAT, we evaluate our re-implementation of FEAT denoted by $\text{FEAT}^*$ with $l\in\{7,13\}$. We maintain equivalent settings in the design of the latent mapper and attention network and only intend to compare the single-layer FEAT-style blending with our multi-layer feedforwarded blending (see \Figureref{img2}). 

\subsection{Results}
\label{subsec:results}
\vspace{-1mm}
\noindent {\bf Merits:}
In \Figureref{compare}, we observe that both the StyleCLIP mapper method and StyleMC result in undesirable edits, such as irrelevant edits to the background. StyleCLIP also reduces the age in the first row and affects the neck region. In the fourth row, we see that in addition to applying the prompt {\it surprised}, it discards the white shirt. StyleMC affects the first three rows' complexion, facial expression, and hair color. As also seen in \cite{feat}, we find that for finer edits (row 1), FEAT-style blending at layer 13 ($\text{FEAT}^*_{13}$) is preferable as also with $\text{FEAT}^*_7$ for coarse edits (rows 2-4). We find that {\it blue eyes} results in unwanted edits when blended at $l=7$, and so does {\it mohawk hairstyle} at $l=13$. 
\begin{table}[t]
    \begin{center}
    \caption{Average Clean-FID~\cite{cleanfid} and training time to desirable quality. Text-prompt legend: {\bf T1)} Happy; {\bf T2)} Surprised; {\bf T3)} Blue eyes; {\bf T4)} Mohawk hairstyle. Method legend: {\bf 1)} SS Global; {\bf 2)} SS Mapper; {\bf 3)} AttnNet-Global; {\bf 4)} AttnNet-Mapper; {\bf 5)} StyleCLIP; {\bf 6)} StyleMC; {\bf 7)} $\text{FEAT}^*$}
    \label{tab:quant_comparison}
    \vspace{-2.5mm}
     \scalebox{0.79}{
    \begin{tabular}{|cc||c|c|c|c|c||c|}
    \hline
    & &\multicolumn{5}{|c|}{Clean-FID ($\downarrow$)}&\multirow{2}{*}{Avg. Time} \\
         \cline{3-7}
& & T1 & T2 & T3 & T4 & Avg. & \\ 
        \hline
          \parbox[t]{0mm}{\multirow{4}{*}{\rotatebox[origin=c]{90}{CORAL}}} 
         & 1 & 4.23 & 4.42& 6.73 & 6.05 &5.35&15min\\
        & 2 & 3.75& 9.28 & 1.75 & 2.73 &4.38&42min\\
        & 3 & 2.22 & 2.26 & 3.08 & 7.20 &3.69&1.2hr\\
        & 4 & 1.85 & 6.38 & 1.38 & 5.29 &3.73&2hr\\
         \cline{1-8}
       \parbox[t]{0mm}{\multirow{3}{*}{\rotatebox[origin=c]{90}{others}}} & 5 &6.98 &18.14 &5.40 &22.50 &13.26&1.5hr\\
        & 6 & 2.93 & 25.12 & 22.15 & 11.30 & 15.38 & 20s \\
        & 7 &  2.51 &  9.46  & 1.93 &  3.01 &  4.23   & 1.8hr \\
         \hline
    \end{tabular}
    }    
    \end{center}
    \vspace{-1cm}
    
\end{table}    

CORAL, however (last four columns in \Figureref{compare}), only affects the relevant regions of interest, which would be the hair region for {\it long curly hair} and {\it mohawk hairstyle}, the eyes and mouth for {\it blue eyes} as well as {\it surprised}. 
These traits persist in \Figureref{teaser}, and \Figureref{67}-A to D wherein the edits are incorporated such that the editing area is minimal and is limited to only the relevant layers. CORAL learns the layers and regions to edit automatically with no domain knowledge or repeated trials. The edits are highly accurate. For example, the prompt {\it mustache} does not also affect the beard, as is apparent from the corresponding masks. 

Under the minimality constraints given by \eqref{eq:wreg_1} and \eqref{eq:wreg_2}, we observe that for enabling finer edits such as {\it blue eyes} and {\it purple hair}, only the latter higher resolution layers (typically $l\geq 8$) are selected, whereas, for coarser structural edits, the earlier smaller layers (typically $l\leq 8$) are automatically selected. 
We clearly see that when CORAL is trained for complex multi-faceted prompts such as {\it curly hair and lipstick} (see \Figureref{complexprompt} and \Figureref{67}-H), the hair edits come from earlier layers whereas the lip edits come from last layers. Furthermore, for such prompts, we found that FEAT blending fails to preserve {\it realism} by introducing noise artifacts (see the example for $\text{FEAT}^*_{13}$ under {\it makeup and afro hairstyle} in \Figureref{complexprompt}). This is also seen in \Figureref{compare} for {\it mohawk} using $\text{FEAT}^*_{13}$.

From \Tableref{quant_comparison}, we see that while the Clean-FID~\cite{cleanfid} of all our edits remains within acceptable bounds of the initially generated distribution, the time required to train CORAL to a desirable edit quality increases with the complexity of the region, layer selector, and the latent editor combined, from method 1 to 4. Segment-selection-based CORAL is significantly faster to train than the attention network. 

We also observe that, on average, segment selection has a higher FID than attention network. Along similar lines, {\it global} edits have a higher FID than {\it latent mapper}, except for {\it surprised}, which we attribute to {\it global} edits predominantly affecting the eyes for this prompt, even for StyleMC, unlike the mapper method which also opens up the mouth.

\noindent {\bf Limitations:}
The segment-selection-based approach trains at a fraction of the time taken by its counterparts, as seen in \Tableref{quant_comparison}. However, the defined segments of a pre-trained segmentation model can affect performance.
For example, in \Figureref{67}-F, our semantic segmentation model combines all the eye and mouth regions into a single semantic segment. As a result, the text prompt {\it lipstick} brightens the skin tone and removes wrinkles from around the eyes. Alternatively, in \Figureref{67}-E, a different segmentation network with dedicated classes for lips overcomes this issue. 

We also note that the quality of the mustache is superior in \Figureref{67}-B compared to A. It turns out that unlike our non-linear mapper which succeeds, the {\it global} edits 
result in black coloration in the {\it mustache} region in many examples. 


%% file: conclusion.tex
CoralStyleCLIP leverages StyleGAN2 and CLIP models to co-optimize region and layer selection for performing high-fidelity text-driven edits on photo-realistic generated images. We demonstrate the efficacy of our generic multi-layer feature blending strategy across varying complexities of the latent editors and region selectors, addressing limitations regarding manual intervention, training complexity, and over- and under-selection of regions along the way. The CORAL strategy can also enhance interactive editing experience by utilizing the predicted masks at each layer.

%% file: notation.tex
$X\sim\mathcal{N}(\mu,\Sigma)$ denotes a Gaussian random vector $X$ with mean $\mu$ and covariance $\Sigma$, and $\mathbf{I}$ denotes the identity matrix. Out of the four well known latent spaces of the StyleGAN2~\cite{sg2}, denoted by $\mathcal{Z}, \mathcal{W}, \mathcal{W}^+$ and $\mathcal{S}$ or the {\it StyleSpace}, CORAL extensively utilizes the $\mathcal{W}^+$ space. MLP abbreviates multi-layer perceptron in this paper. 

For simplicity, $f^{(l)}$ is used to denote original features while $f^{*(l)}$ denotes edited features at each convolutional layer of StyleGAN2. Correspondingly, $I$ denotes the original image and $I^*$ denotes the edited image. $\langle\mathbf{x},\mathbf{y}\rangle$ represents the dot product between vectors \bx and \by. $\|\bx\|_2$ is the Euclidean norm of vector \bx. For any score, $\uparrow$ is used to denote that a higher value is more desirable. The definition for $\downarrow$ follows along similar lines.

%% file: pseudocode.tex
\noindent{\bf StyleGAN2 (\Subsectionref{background}).} We will first present an overview of the StyleGAN2~\cite{sg2} architecture which can be modularized into three parts.
\begin{enumerate}[wide, labelwidth=!, labelindent=0pt]
    \item Mapper $\underset{\mathcal{Z}\rightarrow\mathcal{W}^+}{\text{MLP}}(\cdot)$ from $z\in\R^d$ in $\mathcal{Z}$ space to $\mathbf{w}\coloneqq[w^{(1)},w^{(2)},\ldots,w^{(18)}]\in\R^{18\times d}$ in $\mathcal{W}^+$ space,
    \item 18 learned convolutional blocks $\Phi^{(l)}$ where $l\in\{1,2,\ldots,18\}$, used as $f^{(l)} = \Phi^{(l)} (f^{(l-1)}, w^{(l)})$ emitting features $f^{(l)}\in \R^{H_l\times W_l \times d_l}$ as outputs. Here $H_l\times W_l$ is the resolution at which the features are generated at layer $l$. In addition, a fixed pre-trained tensor $f^{(0)}\coloneqq c\in\R^{4\times 4}$ is learned when training the StyleGAN2 backbone network on a dataset. 
    
    The progressive nature of StyleGAN2 architecture implies that $H_l\leq H_{l+1}$ and $W_l\leq W_{l+1}$. In our experiments, we also have $H_l = W_l$. Furthermore, layers with smaller $l$ synthesize coarser attributes, while the latter layers are seen to control finer attributes, as evident in multiple figures in our paper. 
    \item RGB image constructed as $I=\sum_{l\in\widetilde{L}}RGB^{(l)}(f^{(l)})$ where $I\in\R^{H\times W\times 3}$ and $\widetilde{L}\coloneqq\{2,4,6,\ldots,18\}$. 
\end{enumerate}
Given the latent code $z\in \N(\mathbf{0}, \mathbf{I})$ for a particular image $I$, we can obtain the corresponding $w^{(l)}$ vectors by, $$[w^{(1)},w^{(2)},\ldots,w^{(18)}] = \underset{\mathcal{Z}\rightarrow\mathcal{W}^+}{\text{MLP}}(z)$$ after which the forward pass of the StyleGAN2 generator is given by \Algorithmref{coral1}.

\begin{algorithm}[t]
  \caption{StyleGAN2 forward pass}
  \label{algo:coral1}
  \hspace*{\algorithmicindent} \textbf{Input} $\{w^{(l)}\}_{l=1}^{18}\in\mathcal{W}^+$ space\\
  \hspace*{\algorithmicindent} \textbf{Output} Generated image $I$, features $\{f^{(l)}\}_{l=1}^{18}$ at every layer 

  \begin{algorithmic}[1]
  \Function{Forward}{$\mathbf{w}$}
  \State Set $f^{(0)}= c$ 
  \For{$l\in\{1,2,\ldots,18\}$}
  \State $f^{(l)} = \Phi^{(l)} (f^{(l-1)}, w^{(l)})$
  \If{$l\in\widetilde{L}$}
  \State $I^{(l)} = RGB^{(l)}(f^{(l)})$
  \EndIf
  \EndFor
  \State $I=\sum_{l\in \widetilde{L}}I^{(l)}$ 
  \State \Return $I,\{f^{(l)}\}_{l=1}^{18}$
   
    \EndFunction
    \algstore{myalg}
  \end{algorithmic}

\end{algorithm}

\vspace{2mm}
\noindent{\bf Multi-layer feedforwarded blending (\Subsectionref{blending}).} Alternatively, if instead we have $\mathbf{w_1}\coloneqq \{w_1^{(l)}\}_{l=1}^{18}\in\mathcal{W}^+$,$\mathbf{w_2}\coloneqq\{w_2^{(l)}\}_{l=1}^{18}\in\mathcal{W}^+$, obtained as 
$$\mathbf{w_1}=\underset{\mathcal{Z}\rightarrow\mathcal{W}^+}{\text{MLP}}(z_1) ~~~ \mathbf{w_2}=\underset{\mathcal{Z}\rightarrow\mathcal{W}^+}{\text{MLP}}(z_2)$$
corresponding to two images, a blended forward pass can be performed as described in \Algorithmref{coral2}. Note that the blended forward pass is a sophisticated non-linear spatial interpolation mechanism between the two images generated by $z_1$ and $z_2$. 

\begin{algorithm}[t]
  \caption{StyleGAN2 blended forward pass}
  \label{algo:coral2}
  \hspace*{\algorithmicindent} \textbf{Input} $\mathbf{w_1}, \mathbf{w_2}\in\mathcal{W}^+$, blending masks $m^{(l)}\in [0,1]^{H_l\times W_l}\forall \,l\in\{1,2,\ldots,18\}$\\
  \hspace*{\algorithmicindent} \textbf{Output} Generated image $I$
  \begin{algorithmic}[1]
  \algrestore{myalg}
  \Function{BlendedForward}{$\mathbf{w_1}, \mathbf{w_2}, \{m^{(l)}\}_{l=1}^{18}$}
  \State $\Delta^{(l)}=w_2^{(l)}-w_1^{(l)}$
  \State Set $f^{*(0)}= c$ 
  \For{$l\in\{1,2,\ldots,18\}$}
  \State $\widehat{f^{*(l)}} =  \Phi^{(l)}(f^{*(l-1)}, w_1^{(l)}+\Delta^{(l)})$
  \State $\widehat{f^{(l)}} =  \Phi^{(l)}(f^{*(l-1)}, w_1^{(l)})$
  \State $f^{*(l)} =m^{(l)} \odot \widehat{f^{*(l)}} + (1-m^{(l)}) \odot  \widehat{f^{(l)}}$
  \If{$l\in\widetilde{L}$}
  \State $I^{*(l)} = RGB^{(l)}(f^{*(l)})$
  \EndIf
  \EndFor
  \State $I^*=\sum_{l\in \widetilde{L}}I^{*(l)}$ 
  \State \Return $I^*$
    \EndFunction
    \algstore{myalg}
  \end{algorithmic}

\end{algorithm}

\vspace{2mm}

\noindent{\bf CORAL (\Subsectionref{coral}).} Under the assumption that we can make use of two trainable modules $\textsc{SegmentSelector}(\cdot)$ and $\textsc{ConvAttnNetwork}(\cdot)$ described for segment-selection-based and attention network-based CORAL respectively in \Subsectionref{coral}, we can compute the loss functions in both settings. In the case of segment selection, we use a pre-trained frozen segmentation network and a matrix $e$. The matrix $e \in [0, 1]^{P\times 18}$ is used to model the selection strength of each image segment, where $P$ is the number of semantic segments predicted by the pre-trained segmentation network. In the case of the attention network, we employ a convolution network at each layer of StyleGAN2. More details about the network architecture are described in \Appendixref{arch}. 

\begin{algorithm}[t]
  \caption{CORAL based on segment-selection}
  \label{algo:coral3}
  \hspace*{\algorithmicindent} \textbf{Input} $z\in \N(\mathbf{0}, \mathbf{I})$ in $\mathcal{Z}$ space, text prompt $t$\\
  \hspace*{\algorithmicindent} \textbf{Output} Loss $\mathcal{L}$

  \begin{algorithmic}[1]
  \algrestore{myalg}
  \Function{Loss}{$z,t$}
  \State $\mathbf{w}\coloneqq\{w^{(l)}\}_{l=1}^{18}=\underset{\mathcal{Z}\rightarrow\mathcal{W}^+}{\text{MLP}}(z)$
  \State $\Delta\coloneqq\{\Delta^{(l)}\}_{l=1}^{18}=g(\mathbf{w})$
  \State $\mathbf{w_1}=\mathbf{w},\mathbf{w_2}=\mathbf{w}+\Delta$
  \State $I,\{f^{(l)}\}_{l=1}^{18}=\textsc{Forward}(\mathbf{w_1})$
  \State $\widetilde{I},\{\widetilde{f}^{(l)}\}_{l=1}^{18}=\textsc{Forward}(\mathbf{w_2})$
  \State $e,\{m^{(l)}\}_{l=1}^{18} = \textsc{SegmentSelector}(I)$
  \State $I^*=\textsc{BlendedForward}(\mathbf{w_1}, \mathbf{w_2}, \{m^{(l)}\}_{l=1}^{18})$
  
  \State Loss $\mathcal{L} = \mathcal{L}_{ss}(I,I^*,\widetilde{I},e, \Delta, t)$
  \State \Return $\mathcal{L}$
    \EndFunction
    \algstore{myalg}
  \end{algorithmic}

\end{algorithm}
The function $g(\mathbf{w})$ is used to represent the latent edit direction which could either be a {\it global direction} or an output from a non-linear {\it latent mapper}.
\begin{algorithm}[t]
  \caption{CORAL based on attention network}
  \label{algo:coral4}
  \hspace*{\algorithmicindent} \textbf{Input} $z\in \N(\mathbf{0}, \mathbf{I})$ in $\mathcal{Z}$ space, text prompt $t$\\
  \hspace*{\algorithmicindent} \textbf{Output} Loss $\mathcal{L}$

  \begin{algorithmic}[1]
  \algrestore{myalg}
  \Function{Loss}{$z,t$}
  \State $\mathbf{w}\coloneqq\{w^{(l)}\}_{l=1}^{18}=\underset{\mathcal{Z}\rightarrow\mathcal{W}^+}{\text{MLP}}(z)$
  \State $\Delta\coloneqq\{\Delta^{(l)}\}_{l=1}^{18}=g(\mathbf{w})$
  \State $\mathbf{w_1}=\mathbf{w},\mathbf{w_2}=\mathbf{w}+\Delta$
  \State $I,\{f^{(l)}\}_{l=1}^{18}=\textsc{Forward}(\mathbf{w_1})$
  \State $\widetilde{I},\{\widetilde{f}^{(l)}\}_{l=1}^{18}=\textsc{Forward}(\mathbf{w_2})$
  \State $\{m^{(l)}\}_{l=1}^{18} = \textsc{ConvAttnNetwork}(\{f^{(l)}\}_{l=1}^{18})$
  \State $I^*=\textsc{BlendedForward}(\mathbf{w_1}, \mathbf{w_2}, \{m^{(l)}\}_{l=1}^{18})$
  
  \State Loss $\mathcal{L} = \mathcal{L}_{can}(I,I^*,\widetilde{I},\{m^{(l)}\}_{l=1}^{18}, \Delta, t)$
  \State \Return $\mathcal{L}$
    \EndFunction
  \end{algorithmic}

\end{algorithm}
Finally, the only parameters optimized for minimizing the loss $\mathcal{L}$ are $e$ in segment selection, the parameters of the convolutional attention network for CORAL based on attention network, and finally, those of the latent edit predictor $g(\cdot)$ or $\Delta$ in both variants of CORAL.

%% file: cliploss.tex
\begin{figure}[t]
    \centering
    \includegraphics[width=\linewidth]{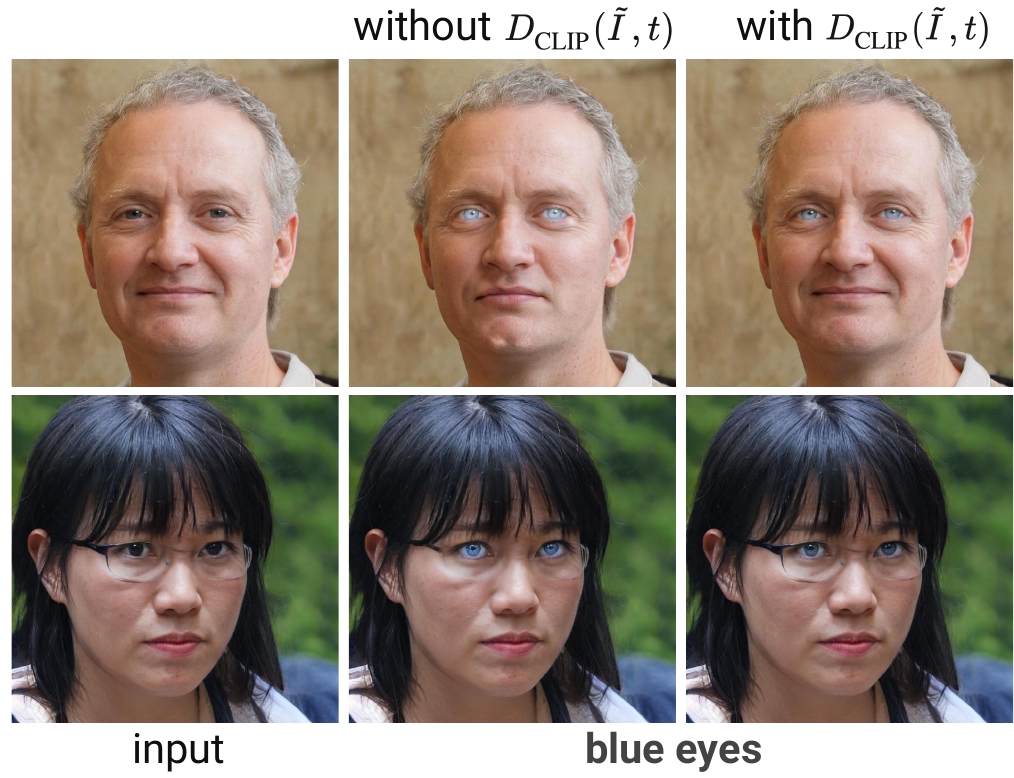}
    \vspace{-2mm}
    \caption{Results with and without additional CLIP loss}
    \vspace{-5mm}
    \label{fig:clip_loss_ablation}
\end{figure}

In \Figureref{clip_loss_ablation}, we see that the additional $D_{\text{CLIP}}(\widetilde{I}, t)$ loss in \eqref{eq:clip} is essential for obtaining high-quality edits. In particular, for the text prompt {\it blue eyes}, we see that without this loss, there are unwanted white patches near the chin in the first row and second column, and the expression of the face is also affected. In the second row, the glasses are removed, and the complexion becomes fairer, thus affecting unrelated attributes. 

With $D_{\text{CLIP}}(\widetilde{I}, t)$, however, the edits are more precise, and only the eye region is affected. This is happening because the image corresponding to $\mathbf{w_2}$ in \Algorithmref{coral2} is also explicitly driven to be semantically aligned with the text prompt $t$ and therefore provides a better reference image for guiding the interpolation.

%% file: architecture_diagram.tex
\begin{figure*}[t]
    \centering
    \includegraphics[width=\linewidth]{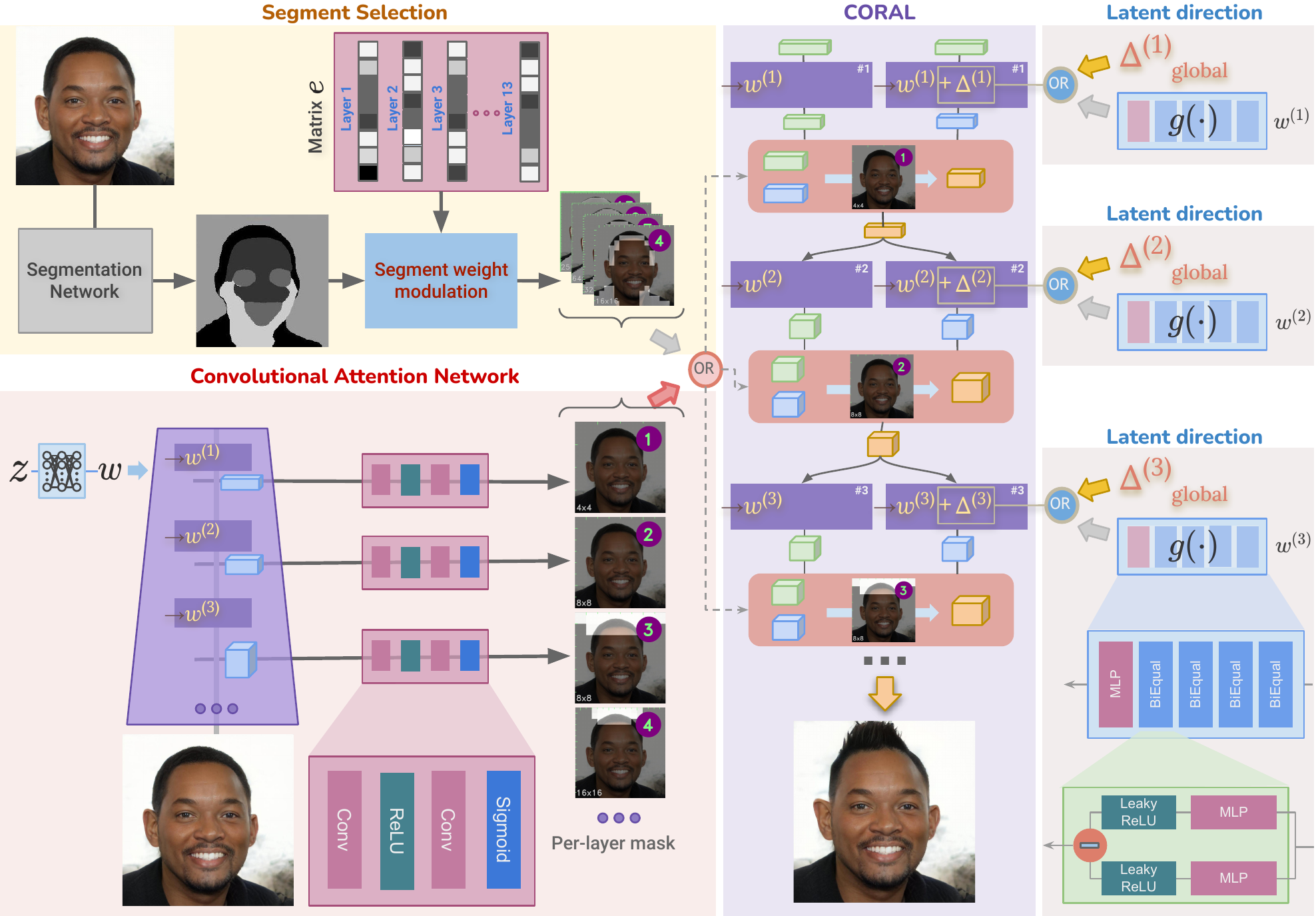}
    \caption{Architecture diagram of CoralStyleCLIP. \textbf{Segment selection network} consists of a pre-trained segmentation network and matrix $e$. The weights of each segment are modulated to produce a CORAL mask. \textbf{Convolutional attention network} consists of a CNN which predicts the CORAL mask at each layer of StyleGAN. The CORAL mask can either come from Segment selection \textit{or} Convolutional Attention Network.
    \textbf{Latent direction} can either come from the learnt global direction $\Delta$ \textit{or} via a mapper $g(\cdot)$ at each layer of CoralStyleCLIP. The layers of CoralStyleCLIP blend the features using the mask and latent direction (See Suppl. Pseudocode). There are three $g(\cdot)$ modules for coarse ($l \in [1, 4]$), medium ($l\in[5, 8]$) and fine layers ($l\in[9, 18]$) each. In this figure, the result of {\it mohawk hairstyle} used a convolutional attention network and global direction.
    }
    \label{fig:suppl_archi}
\end{figure*}

In \Figureref{suppl_archi}, we present the architecture diagram of CoralStyleCLIP with all components. In this paper, we demonstrated CORAL with four different variants. We demonstrated two approaches for predicting CORAL masks - segment selection and convolutional attention network. We also demonstrated results on two different variations of the latent direction - global direction $\Delta$  and latent mapper network $g(\cdot)$. Therefore, in total, we have four combinations of CORAL variants. For segment selection, we have a matrix $e$, which is used to modulate the weights of each image segment. In the case of the Convolutional Attention Network, we employ a CNN at each layer of the StyleGAN network. Architecture details of the network are mentioned in \Sectionref{expmt} of the main paper. 

\noindent \textbf{Latent Mapper}: As also discussed in \Subsectionref{lated}, the latent mapper $g(\cdot)$ is an MLP-based model along the lines of \cite[Section 5]{styleclip}, where the $w^{(l)}$ are split into three groups: coarse ($l$ in 1 to 4), medium ($l$ in 5 to 8) and fine ($l$ in 9 to 18); and each of these groups is processed by a different MLP.
Our latent mapper network consists of four BiEqual linear layers followed by a multi-layer perceptron. Each BiEqual layer consists of two MLPs followed by a LeakyReLU~\cite{leakyrelu} activation function and a differencing operation (\Figureref{suppl_archi}).

\begin{table}[t]
    \centering
    \caption{Metrics for Text Prompt - {\it Blue eyes}}
    \vspace{-3mm}
    \label{tab:blueeyes}  
    \resizebox{\columnwidth}{!}{%
    \begin{tabular}{|c|c|c|c|c|}
    \hline
     & ID ($\uparrow$) & LPIPS  ($\downarrow$) & MS-SSIM ($\uparrow$) & $L_2$  ($\downarrow$) \\ 
    \hline
     SS Global & 0.776 & 0.0160 & 0.972 & 0.0029 \\
  
    SS Mapper & 0.799& 0.0082 & 0.969 & 0.0023\\
  
    AttnNet-Global & 0.868 & 0.0067 & 0.988 & 0.0015 \\
  
     AttnNet-Mapper & 0.896 & 0.0060 & 0.989 & 0.0012 \\
     \hline
     StyleCLIP & 0.741 & 0.0856  &  0.871 & 0.0265\\
  
    StyleMC & 0.522 & 0.1670 &  0.596 & 0.2550 \\
  
    $\text{FEAT}^*$ & 0.904 & 0.0017 & 0.997 & 0.0004 \\
    \hline
    \end{tabular}
    }
    \vspace{-2.7mm}
    
\end{table}

\begin{table}[t]
    \centering
    \caption{Metrics for Text Prompt - {\it Happy}}
    \vspace{-3mm}
  \label{tab:happy} 
    \resizebox{\columnwidth}{!}{%
    \begin{tabular}{|c|c|c|c|c|}
    \hline
     & ID ($\uparrow$) & LPIPS  ($\downarrow$) & MS-SSIM ($\uparrow$) & $L_2$  ($\downarrow$) \\ 
    \hline
     SS Global & 0.633 & 0.0313 & 0.935 & 0.0080 \\
  
    SS Mapper & 0.651 & 0.0308 &  0.933 & 0.0084 \\
  
    AttnNet-Global & 0.830 &  0.0136 & 0.961 & 0.0050 \\
  
     AttnNet-Mapper & 0.847 & 0.0155 & 0.959 & 0.0053 \\
     \hline
     StyleCLIP & 0.644 & 0.0904 & 0.835 & 0.0301\\
  
    StyleMC & 0.821 & 0.0244 & 0.940  & 0.0080 \\
  
    $\text{FEAT}^*$ & 0.846 & 0.0150 & 0.957 & 0.0064 \\
    \hline
    \end{tabular}
    }
    \vspace{-2.7mm}
\end{table}

\begin{table}[t]
    \centering
\caption{Metrics for Text Prompt - {\it Mohawk hairstyle}}
\vspace{-3mm}
    \label{tab:mohawk} 
    \resizebox{\columnwidth}{!}{%
    
    \begin{tabular}{|c|c|c|c|c|}
    \hline
     & ID ($\uparrow$) & LPIPS  ($\downarrow$) & MS-SSIM ($\uparrow$) & $L_2$  ($\downarrow$) \\ 
    \hline
     SS Global & 0.828 & 0.1980 & 0.756 & 0.0760 \\
  
    SS Mapper & 0.882 & 0.0970 & 0.868 & 0.0303\\
  
    AttnNet-Global & 0.945 & 0.0849 & 0.919 & 0.0289 \\
  
     AttnNet-Mapper & 0.922 & 0.0971 & 0.899 & 0.0337 \\
     \hline
     StyleCLIP & 0.522 & 0.2940 &  0.597 & 0.1530 \\
  
    StyleMC & 0.651 & 0.0704 & 0.898 & 0.0180 \\
  
    $\text{FEAT}^*$ & 0.953 & 0.0746 & 0.924 & 0.0236 \\
    \hline
    \end{tabular}
    }
    \vspace{-2.7mm}
\end{table}

\begin{table}[t]
    \centering
    \caption{Metrics for Text Prompt - {\it Surprised}}
    \vspace{-3mm}
        \label{tab:surprised}   
    \resizebox{\columnwidth}{!}{%
    
    \begin{tabular}{|c|c|c|c|c|}
    \hline
     & ID ($\uparrow$) & LPIPS  ($\downarrow$) & MS-SSIM ($\uparrow$) & $L_2$  ($\downarrow$) \\ 
    \hline
     SS Global & 0.747 & 0.0173 & 0.972 & 0.0037 \\
  
    SS Mapper & 0.519 & 0.0412 & 0.902 & 0.0122\\
  
    AttnNet-Global & 0.819 & 0.0167 & 0.973 & 0.0035 \\
  
     AttnNet-Mapper & 0.654 & 0.0297 & 0.927 & 0.0092 \\
     \hline
     StyleCLIP & 0.633 & 0.1390 & 0.765 &  0.0496 \\
  
    StyleMC & 0.602 & 0.0952 & 0.728 & 0.0713 \\
  
    $\text{FEAT}^*$ & 0.719 & 0.0252 & 0.938 & 0.0088 \\
    \hline
    \end{tabular}
    }
    \vspace{-2.7mm}
    
\end{table}

%% file: additional_experiment_details.tex
We provide the hyperparameters used for training CoralStyleCLIP for making edits to images generated by a StyleGAN trained on the FFHQ dataset. Note that for both segment selection and attention network, the user can potentially decrease the $\lambda_{id}$ by 20-40\% depending on how likely the prompt is to make any edit to the facial region. This is essential for text prompts such as {\it kid}, {\it elderly}, and {\it asian}, where the transformation can alter the identity of a person significantly. 

\noindent{\bf Segment selection:} For experiments based on global directions (Segment Selection - Global Direction), we set $\lambda_{l_2}=0.0007,\lambda_{id}=0.015,\lambda_{area}=0.10$, whereas for latent mapper based edits (Segment Selection - Mapper), we set $\lambda_{l_2}=0.0002,\lambda_{id}=0.020,\lambda_{area}=0.08$. 

\noindent{\bf Attention network:} For experiments based on global directions (Attention Network - Global Direction) we set $\lambda_{l_2}=0.0009,\lambda_{id}=0.08,\lambda_{area}=0.00009,\lambda_{tv}=0.00003$, whereas for latent mapper based edits (Attention Network - Mapper), we set $\lambda_{l_2}=0.0006,\lambda_{id}=0.08,\lambda_{area}=0.00002,\lambda_{tv}=0.00003$ 

In \Subsectionref{cars_more_edits}, \Subsectionref{sketch_more_edits} and \Subsectionref{pixar_more_edits}, we also present results for attention network-based CORAL with {\it latent mapper} edits. For the Stanford cars dataset, $\lambda_{l_2}$ is set to 0.0002 while keeping other hyperparameters the same. For the remaining two domains, which are adaptations of FFHQ, $\lambda_{l_2}$ is set to 0.0004.

%% file: additional_results.tex
In addition to presenting results for editing images using CoralStyleCLIP on the FFHQ dataset~\cite{sg1} (see \Figureref{ffhq}), we also demonstrate the benefits of our method for the Stanford Cars dataset~\cite{stanfordcars} (see \Figureref{cars}), and for face generators which were adapted to the following domains: {\it sketch}, and {\it pixar} (see \Figureref{sketch} and \Figureref{pixar} respectively), using StyleGAN-NADA\cite{nada}. 

For experiments other than those with FFHQ, we disable the ID loss $\mathcal{L}_{id}$. Nonetheless, we do observe high-fidelity edits in these settings as well. Also note that as mentioned in \Sectionref{expmt}, for $l>13$, we set the masks $m^{(l)}=\mathbf{0}$ for CoralStyleCLIP.

\subsection{FFHQ~\cite{sg1}}
\label{subsec:ffhq_more_edits}
\begin{figure*}
    \centering
    \includegraphics[height=\textheight]{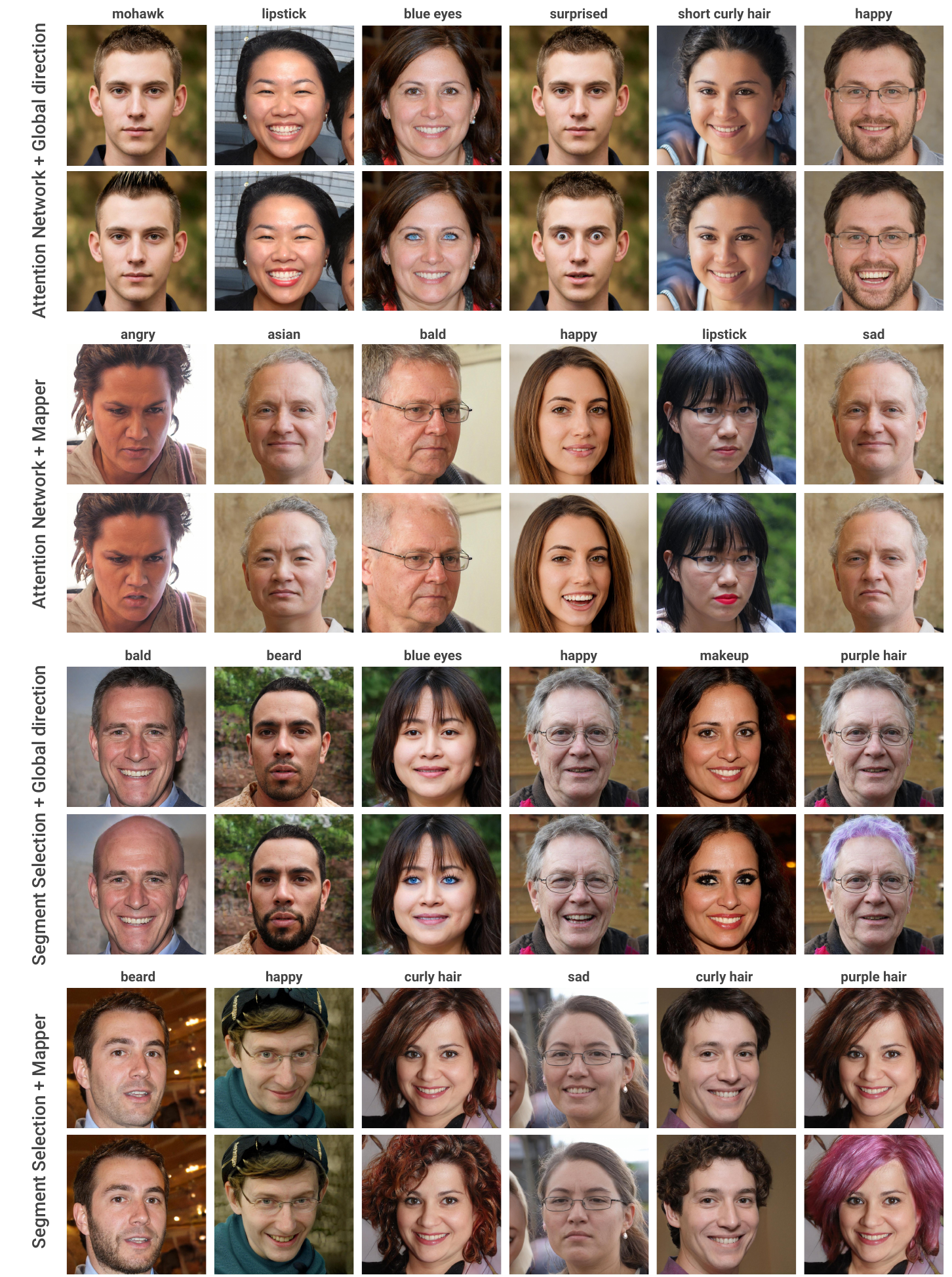}
    \caption{Additional results on FFHQ~\cite{sg1} generated images.}
    \label{fig:ffhq}
\end{figure*}

In \Figureref{ffhq}, we present more examples where CoralStyleCLIP executes a range of edits for human faces with high precision and minimal hand-holding. We choose a variety of text prompts that demand challenging structural and color edits. These results show that our method can accurately select the correct region and layer.

\subsection{Cars~\cite{stanfordcars}}
\label{subsec:cars_more_edits}
\begin{figure*}
    \centering
    \includegraphics[height=\textheight]{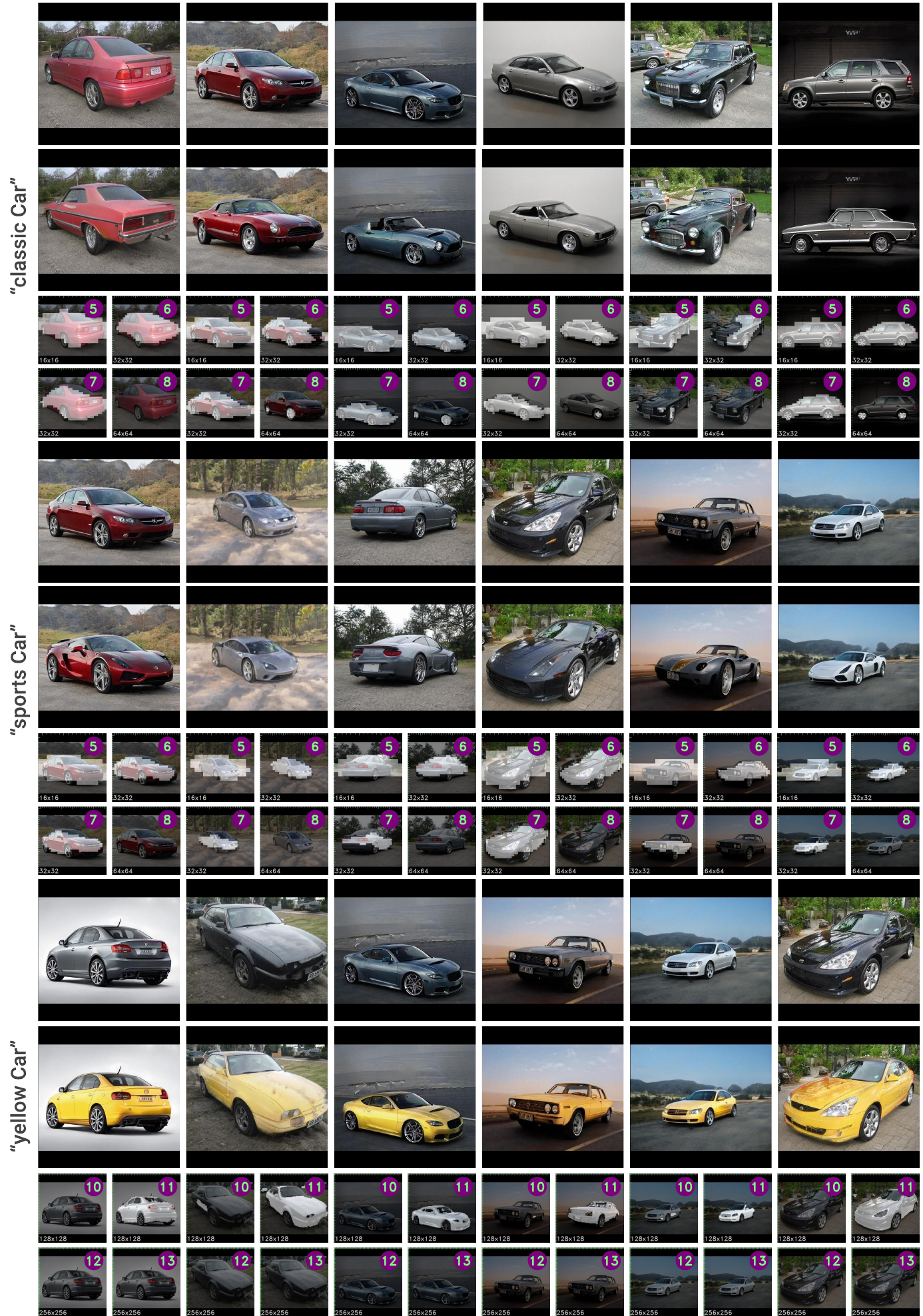}
    \caption{Additional results on Cars~\cite{stanfordcars} generated images.}
    \label{fig:cars}
\end{figure*}

We trained CoralStyleCLIP based on convolutional attention network and the {\it latent mapper} edit for the $512 \times 512$ size StyleGAN2 model trained on Stanford Cars dataset~\cite{stanfordcars} available from \cite{hyperstyle}. For text prompts {\it classic}, {\it sports}, and {\it yellow car}, we observe that only the car is edited while the background is not selected in \Figureref{cars}. Furthermore, we also observe that CoralStyleCLIP automatically selects earlier layers for executing the first two prompts while it utilizes the latter layers to change the car's color. It is also interesting to note that the network generally selects layer 8 for wheel modification while it selects layers 5-7 for editing the car's body. This indicates that the car's wheels will likely have a more significant structural disentanglement and edit flexibility in layer 8. Furthermore, for {\it yellow car}, CORAL prioritizes the car's body over the wheels and windows. We see that the car's color remains preserved for {\it classic car} and {\it sports car}.

\subsection{Sketch~\cite{nada}}
\label{subsec:sketch_more_edits}
\begin{figure*}
    \centering
    \includegraphics[width=\textwidth]{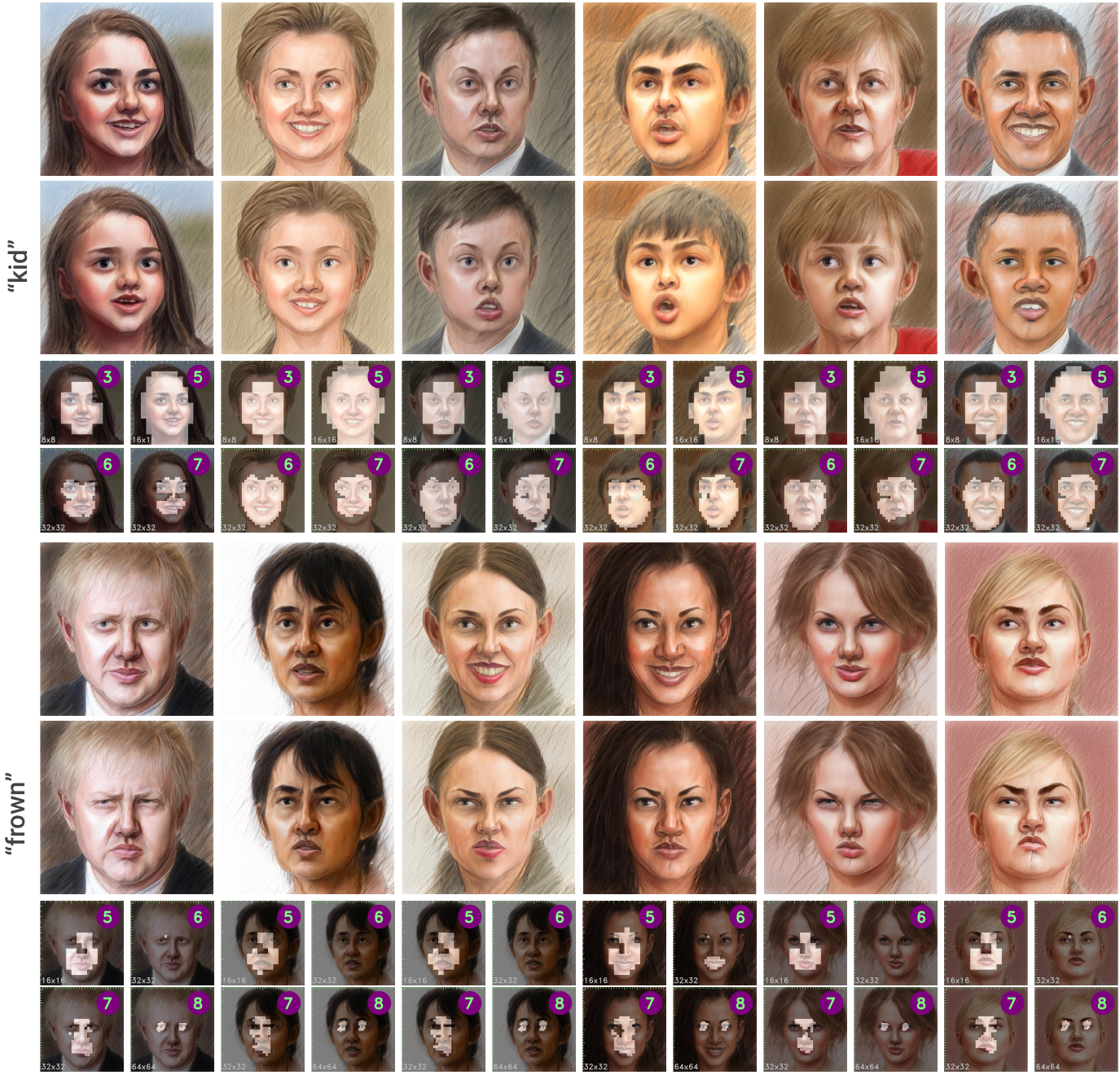}
    \vspace{-3mm}
    \caption{Additional results on Sketch~\cite{nada} generated images.}
    \label{fig:sketch}
    \vspace{-2mm}
\end{figure*}

We trained CoralStyleCLIP based on convolutional attention network and the {\it latent mapper} edit for a StyleGAN that was pre-trained on FFHQ~\cite{sg1}, and then domain adapted to {\it sketches} using StyleGAN-NADA~\cite{nada}. For both prompts {\it kid} and {\it frown}, we see that CORAL successfully identifies the appropriate regions and layers for editing and executes them in \Figureref{sketch}. In the case of {\it kid}, the network selects the facial region, and in {\it frown}, the network selects the eyes and mouth regions. Similar to the observation made in \textit{Cars} subsection above (\Subsectionref{cars_more_edits}), the network selects layer 8 for structural changes to the eyes and selects other layers for making facial edits to achieve  {\it frown}.

\subsection{Pixar~\cite{nada}}
\label{subsec:pixar_more_edits}
\begin{figure*}
    \centering
    \includegraphics[width=\textwidth]{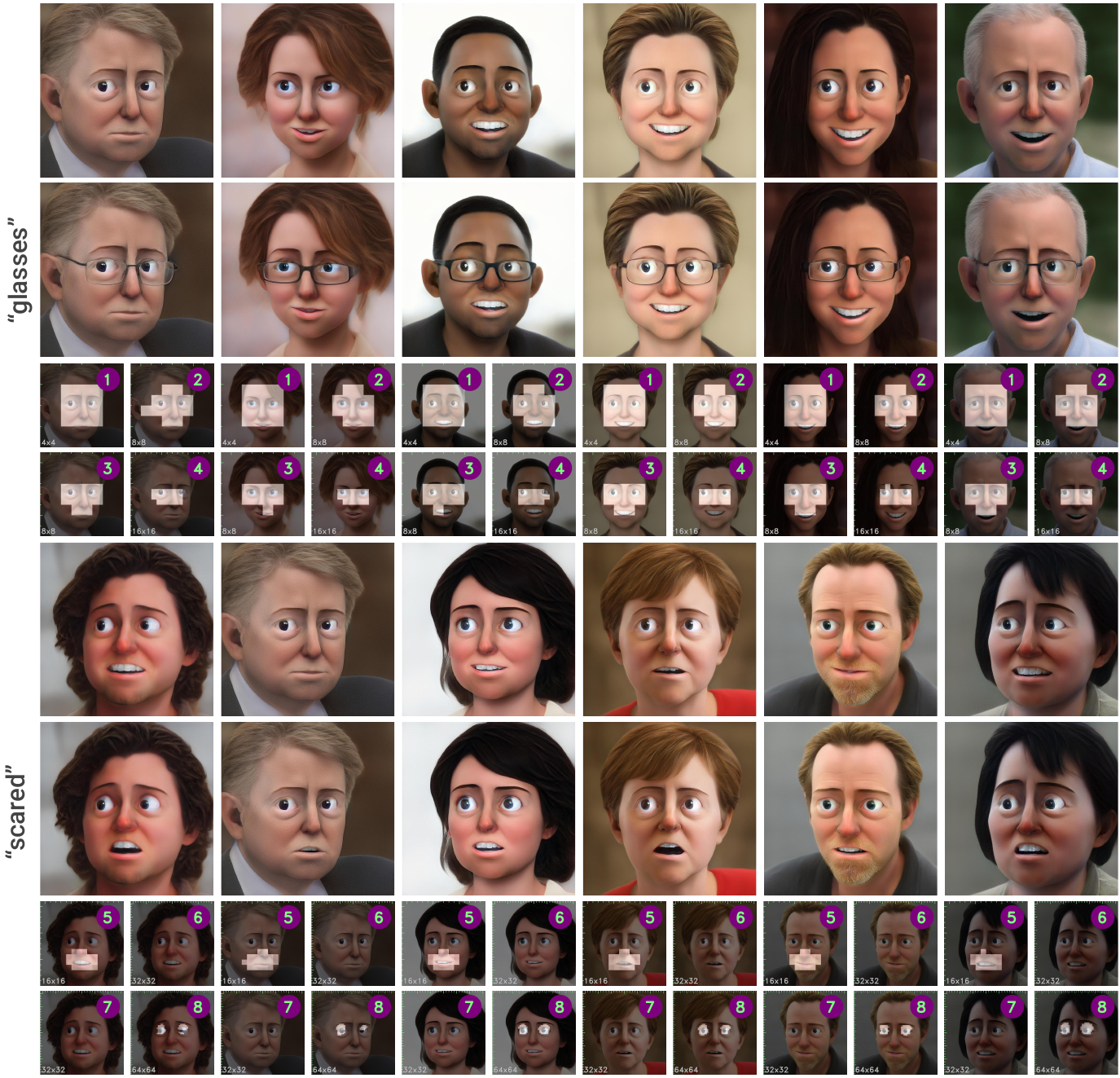}
    \caption{Additional results on Pixar~\cite{nada} generated images.}
    \label{fig:pixar}
    \vspace{-0.5cm}
\end{figure*}

Along the lines of \Subsectionref{sketch_more_edits}, in \Figureref{pixar}, we also apply edits corresponding to {\it glasses} and {\it scared} for StyleGAN2 generated images adapted for the {\it pixar} domain using StyleGAN-NADA~\cite{nada}.

While the {\it glasses} emerge from layers 1 to 4, leaving other attributes undisturbed, {\it scared} affects both the eyes and mouth regions, their corresponding edits emerging from layers 5 and 8, respectively. In both cases, the edits are primarily structural and executed through the earlier layers. These results demonstrate that CoralStyleCLIP can determine the correct regions to edit at the correct set of StyleGAN2 layers with minimal hand-holding.

\subsection{Additional quantitative results (\Tableref{blueeyes} to \ref{tab:surprised})}
\label{subsec:numerics}
In \Tableref{quant_comparison}, we observed that CORAL edits on human faces achieve a reasonable Clean-FID~\cite{cleanfid}, thereby preserving the {\it realism} of the edited images. 

We also compare CORAL for a StyleGAN2 trained on FFHQ with our baselines across the four text prompts from \Tableref{quant_comparison} based on other quantitative metrics. In particular, we compute the identity similarity (ID) based on the cosine similarity between ArcFace embeddings~\cite{arcface}, LPIPS~\cite{lpips} distance, MS-SSIM~\cite{bovik} score and the pixel-wise Euclidean distance, between each edited image and the original. 

For comparison with FEAT~\cite{feat}, we only compare CORAL with results from our reimplementation $\text{FEAT}^*_7$ for {\it happy}, {\it mohawk} and {\it surprised}, and  $\text{FEAT}^*_{13}$ for {\it blue eyes} so that the edits do occur with high precision. On average, the ID is higher in \Tableref{blueeyes} and \Tableref{mohawk} as compared to \Tableref{happy} and \Tableref{surprised}. This is because edits to the facial regions, such as the mouth and eyes, diminish the facial recognition capabilities of ArcFace \cite{arcface}. 

In general, CORAL, based on the attention network, has higher similarity and lower dissimilarity scores than the segment-selection-based methods, which can be attributed to higher precision in the region of interest selection at every layer of the generator. However, the same cannot be said for StyleCLIP and StyleMC, which affect irrelevant regions, consequently underperforming CORAL and FEAT.

%% file: multilayer_feat.tex
\begin{figure}[t]
    \centering
    \includegraphics[width=\linewidth]{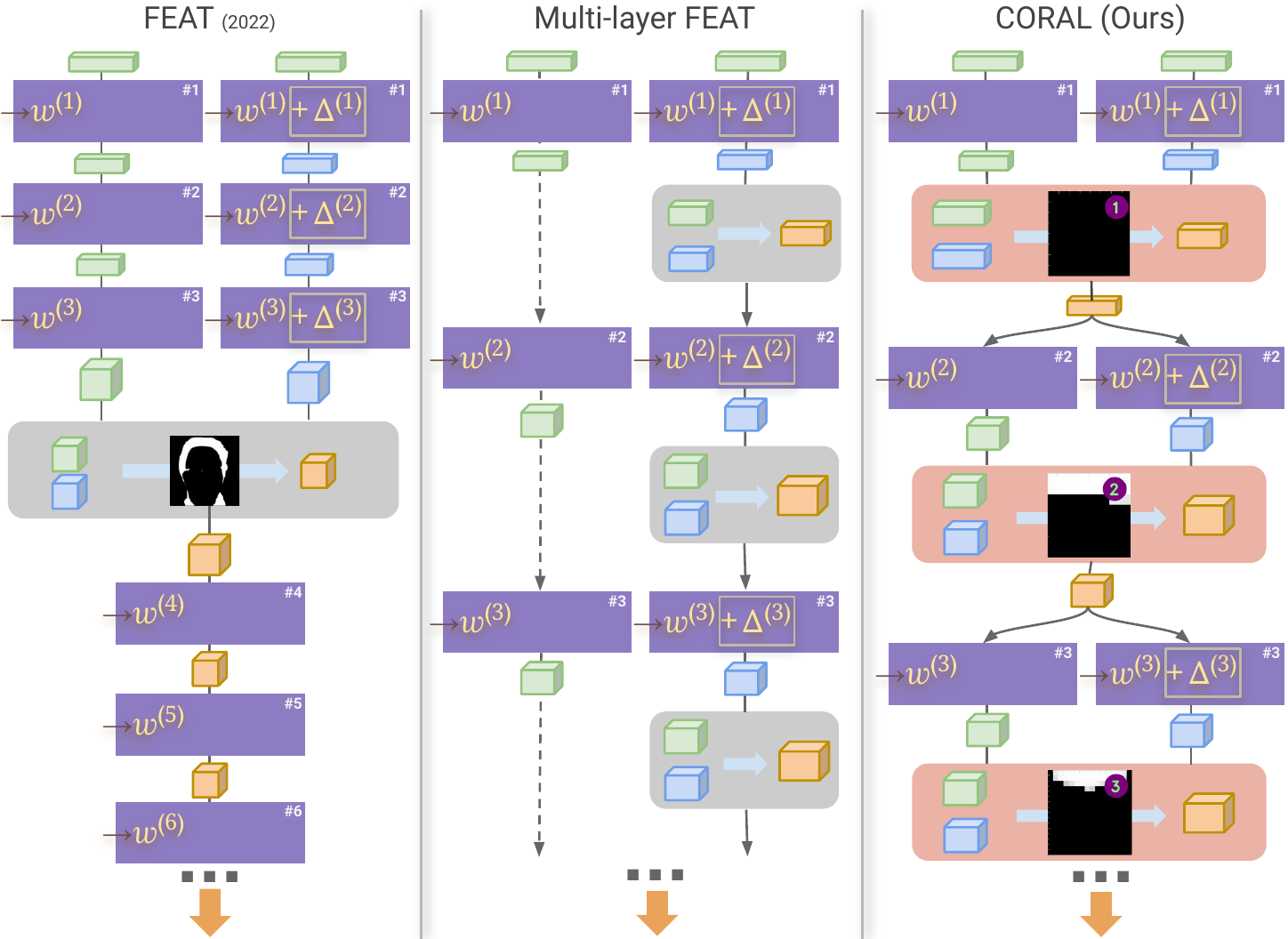}
    \vspace{-6mm}
    \caption{FEAT vs Multi-layer FEAT vs CORAL}
    \label{fig:multi-layer-feat}
    \vspace{-5mm}
\end{figure}

CORAL method applies the proposed feedforwarded feature blending strategy at each of the StyleGAN layers (\Subsectionref{blending}). On similar lines, an interesting and a direct extension of FEAT would be to apply their proposed feature blending strategy at every StyleGAN layer. The resulting method, which we refer to as ``multi-layer FEAT'', would have a similar high level architecture with two parallel pathways but would differ in the blending strategies (\Figureref{multi-layer-feat}). In the case of the multi-layer FEAT, the blended features would be propagated through the edited pathway ($w+\Delta$) while the original features would continue to propagate through the original unedited ($w$) pathway. In other words, the FEAT blending strategy draws relevant edits from a particular layer and learns to discard edits from all the previous layers. Therefore,  multi-layer FEAT cannot propagate the edits effectively. However, in the case of CORAL, blended features are passed parallelly to both pathways (\Subsectionref{blending}). As a result, CORAL can discard irrelevant edits at the current layer and propagate the updated edits effectively.

%% file: impact.tex
\noindent The edits outside the regions of interest are explicitly prevented by the feed-forwarded blending strategy proposed in \Subsectionref{blending}. As a result, methods such as \cite{styleclip,styleflow,stylefusion,stylemc,feat,editgan} could directly benefit from applying our strategy for inference with separately constructed masks. 

Furthermore, the synergy between the co-optimized region and layer selection (CORAL) in \Subsectionref{coral} and the feed-forwarded blending may provide increased control to content designers who may find it easier to refine an accurate initial predicted region of interest.

%% file: main.bbl
\begin{thebibliography}{10}\itemsep=-1pt

\bibitem{age}
Yuval A., Or Patashnik, and Daniel Cohen{-}Or.
\newblock Only a matter of style: age transformation using a style-based
  regression model.
\newblock {\em {ACM} Trans. Graph.}, 40(4):45:1--45:12, 2021.

\bibitem{im2style}
Rameen Abdal, Yipeng Qin, and Peter Wonka.
\newblock Image2stylegan: How to embed images into the stylegan latent space?
\newblock In {\em ICCV}, 2019.

\bibitem{clip2stylegan}
Rameen Abdal, Peihao Zhu, John Femiani, Niloy~J. Mitra, and Peter Wonka.
\newblock Clip2stylegan: Unsupervised extraction of stylegan edit directions.
\newblock In {\em SIGGRAPH}. {ACM}, 2022.

\bibitem{styleflow}
Rameen Abdal, Peihao Zhu, Niloy~J. M., and Peter Wonka.
\newblock Styleflow: Attribute-conditioned exploration of stylegan-generated
  images using conditional continuous normalizing flows.
\newblock {\em {ACM} Trans. Graph.}, 40(3):21:1--21:21, 2021.

\bibitem{video}
Yuval Alaluf, Or Patashnik, Zongze Wu, Asif Zamir, Eli Shechtman, Dani
  Lischinski, and Daniel Cohen{-}Or.
\newblock Third time's the charm? image and video editing with stylegan3.
\newblock {\em ECCVw}, 2022.

\bibitem{hyperstyle}
Yuval Alaluf, Omer Tov, Ron Mokady, Rinon Gal, and Amit Bermano.
\newblock Hyperstyle: Stylegan inversion with hypernetworks for real image
  editing.
\newblock In {\em CVPR}, 2022.

\bibitem{blenddiff}
Omri Avrahami, Dani Lischinski, and Ohad Fried.
\newblock Blended diffusion for text-driven editing of natural images.
\newblock In {\em CVPR}, 2022.

\bibitem{stylegan_survey}
Amit~H. B., Rinon Gal, Yuval Alaluf, Ron Mokady, Yotam Nitzan, Omer Tov, Or
  Patashnik, and Daniel Cohen{-}Or.
\newblock State-of-the-art in the architecture, methods and applications of
  stylegan.
\newblock {\em Comp. Graph. Forum}, 41(2):591--611, 2022.

\bibitem{biggan}
Andrew Brock, Jeff Donahue, and Karen Simonyan.
\newblock Large scale {GAN} training for high fidelity natural image synthesis.
\newblock In {\em ICLR}, 2019.

\bibitem{collins2020editing}
Edo Collins, Raja Bala, Bob Price, and Sabine Süsstrunk.
\newblock Editing in style: Uncovering the local semantics of {GANs}.
\newblock {\em CVPR}, 2020.

\bibitem{arcface}
Jiankang Deng, Jia Guo, Jing Yang, Niannan Xue, Irene Kotsia, and Stefanos
  Zafeiriou.
\newblock Arcface: Additive angular margin loss for deep face recognition.
\newblock {\em {IEEE} TPAMI}, 2022.

\bibitem{Dong2017SemanticIS}
H. Dong, Simiao Yu, Chao Wu, and Y. Guo.
\newblock Semantic image synthesis via adversarial learning.
\newblock {\em ICCV}, pages 5707--5715, 2017.

\bibitem{nada}
Rinon Gal, Or Patashnik, Haggai Maron, Amit~H. Bermano, Gal Chechik, and Daniel
  Cohen{-}Or.
\newblock Stylegan-nada: Clip-guided domain adaptation of image generators.
\newblock {\em {ACM} Trans. Graph.}, 41(4):141:1--141:13, 2022.

\bibitem{gan}
Ian~J. Goodfellow, Jean Pouget{-}Abadie, Mehdi M., Bing Xu, David
  Warde{-}Farley, Sherjil Ozair, Aaron C., and Yoshua Bengio.
\newblock Generative adversarial nets.
\newblock In {\em NeurIPS}, 2014.

\bibitem{ganspace}
Erik H{\"{a}}rk{\"{o}}nen, Aaron Hertzmann, Jaakko Lehtinen, and Sylvain Paris.
\newblock Ganspace: Discovering interpretable {GAN} controls.
\newblock In {\em NeurIPS}, 2020.

\bibitem{feat}
Xianxu Hou, Linlin Shen, Or Patashnik, Daniel Cohen{-}Or, and Hui Huang.
\newblock {FEAT:} face editing with attention.
\newblock {\em CoRR}, abs/2202.02713, 2022.

\bibitem{stylefusion}
Omer Kafri, Or Patashnik, Yuval A., and Daniel Cohen-Or.
\newblock Stylefusion: Disentangling spatial segments in stylegan-generated
  images.
\newblock {\em ACM Trans. Graph.}, Mar 2022.

\bibitem{progan}
Tero Karras, Timo Aila, Samuli Laine, and Jaakko Lehtinen.
\newblock Progressive growing of gans for improved quality, stability, and
  variation.
\newblock In {\em ICLR}, 2018.

\bibitem{sg1}
Tero Karras, Samuli Laine, and Timo Aila.
\newblock A style-based generator architecture for generative adversarial
  networks.
\newblock {\em {IEEE} TPAMI}, 43(12):4217--4228, 2021.

\bibitem{sg2}
Tero Karras, Samuli Laine, Miika Aittala, Janne Hellsten, Jaakko Lehtinen, and
  Timo Aila.
\newblock Analyzing and improving the image quality of stylegan.
\newblock In {\em CVPR}, 2020.

\bibitem{stylemapgan}
Hyunsu Kim, Yunjey Choi, Junho Kim, Sungjoo Yoo, and Youngjung Uh.
\newblock Exploiting spatial dimensions of latent in {GAN} for real-time image
  editing.
\newblock In {\em CVPR}, 2021.

\bibitem{adam}
Diederik~P. Kingma and Jimmy Ba.
\newblock Adam: {A} method for stochastic optimization.
\newblock In Yoshua Bengio and Yann LeCun, editors, {\em ICLR}, 2015.

\bibitem{stylemc}
Umut Kocasari, Alara Dirik, Mert Tiftikci, and Pinar Yanardag.
\newblock Stylemc: Multi-channel based fast text-guided image generation and
  manipulation.
\newblock In {\em WACV}, 2022.

\bibitem{stanfordcars}
Jonathan Krause, Michael Stark, Jia Deng, and Li Fei{-}Fei.
\newblock 3d object representations for fine-grained categorization.
\newblock In {\em 2013 {IEEE} International Conference on Computer Vision
  Workshops, {ICCV} Workshops 2013, Sydney, Australia, December 1-8, 2013},
  pages 554--561. {IEEE} Computer Society, 2013.

\bibitem{Lee2022SoundGuidedSI}
Seung~H. L., Won~Kyoung Roh, Wonmin Byeon, Sang~Ho Yoon, Chan~Y. K., Jinkyu
  Kim, and Sangpil Kim.
\newblock Sound-guided semantic image manipulation.
\newblock {\em CVPR}, 2022.

\bibitem{manigan}
Bowen Li, Xiaojuan Qi, Thomas Lukasiewicz, and Philip H.~S. Torr.
\newblock Manigan: Text-guided image manipulation.
\newblock In {\em CVPR}, 2020.

\bibitem{editgan}
Huan Ling, Karsten Kreis, Daiqing Li, Seung~Wook Kim, Antonio Torralba, and
  Sanja Fidler.
\newblock Editgan: High-precision semantic image editing.
\newblock In Marc'Aurelio Ranzato, Alina Beygelzimer, Yann~N. Dauphin, Percy
  Liang, and Jennifer~Wortman Vaughan, editors, {\em Advances in Neural
  Information Processing Systems 34: Annual Conference on Neural Information
  Processing Systems 2021, NeurIPS 2021, December 6-14, 2021, virtual}, pages
  16331--16345, 2021.

\bibitem{Liu2020DescribeWT}
Yahui Liu, Marco~De Nadai, Deng Cai, Huayang Li, Xavier Alameda-Pineda, N.
  Sebe, and Bruno Lepri.
\newblock Describe what to change: A text-guided unsupervised image-to-image
  translation approach.
\newblock {\em ACMMM}, 2020.

\bibitem{relu}
Vinod Nair and Geoffrey~E. Hinton.
\newblock Rectified linear units improve restricted boltzmann machines.
\newblock In Johannes F{\"{u}}rnkranz and Thorsten Joachims, editors, {\em
  ICML}, 2010.

\bibitem{Nam2018TextAdaptiveGA}
Seonghyeon Nam, Yunji Kim, and S. Kim.
\newblock Text-adaptive generative adversarial networks: Manipulating images
  with natural language.
\newblock In {\em NeurIPS}, 2018.

\bibitem{glide}
Alexander~Quinn Nichol, Prafulla Dhariwal, Aditya Ramesh, Pranav Shyam, Pamela
  Mishkin, Bob McGrew, Ilya Sutskever, and Mark Chen.
\newblock {GLIDE:} towards photorealistic image generation and editing with
  text-guided diffusion models.
\newblock In {\em ICML}, 2022.

\bibitem{anima2020disentangled}
Weili Nie, Tero Karras, Animesh Garg, Shoubhik Debnath, Anjul Patney, Ankit~B.
  Patel, and Animashree Anandkumar.
\newblock Semi-supervised stylegan for disentanglement learning.
\newblock In {\em ICML}, 2020.

\bibitem{unsupseg}
Daniil Pakhomov, Sanchit Hira, Narayani Wagle, Kemar~E. Green, and Nassir
  Navab.
\newblock Segmentation in style: Unsupervised semantic image segmentation with
  stylegan and {CLIP}.
\newblock {\em CoRR}, abs/2107.12518, 2021.

\bibitem{SAM}
Gaurav Parmar, Yijun Li, Jingwan Lu, Richard Zhang, Jun{-}Yan Zhu, and
  Krishna~Kumar Singh.
\newblock Spatially-adaptive multilayer selection for {GAN} inversion and
  editing.
\newblock In {\em CVPR}, 2022.

\bibitem{cleanfid}
Gaurav Parmar, Richard Zhang, and Jun{-}Yan Zhu.
\newblock On aliased resizing and surprising subtleties in {GAN} evaluation.
\newblock In {\em CVPR}, 2022.

\bibitem{styleclip}
Or Patashnik, Zongze Wu, Eli Shechtman, Daniel Cohen{-}Or, and Dani Lischinski.
\newblock Styleclip: Text-driven manipulation of stylegan imagery.
\newblock In {\em ICCV}, 2021.

\bibitem{clip2latent}
Justin N.~M. Pinkney and Chuan Li.
\newblock clip2latent: Text driven sampling of a pre-trained stylegan using
  denoising diffusion and {CLIP}.
\newblock {\em BMVC}, 2022.

\bibitem{clip}
Alec Radford, Jong~W. K., Chris Hallacy, Aditya Ramesh, Gabriel Goh, Sandhini
  Agarwal, Girish Sastry, Amanda Askell, Pamela Mishkin, Jack Clark, Gretchen
  Krueger, and Ilya Sutskever.
\newblock Learning transferable visual models from natural language
  supervision.
\newblock In {\em ICML}, 2021.

\bibitem{ramesh_hierarchy}
Aditya Ramesh, Prafulla Dhariwal, Alex Nichol, Casey Chu, and Mark Chen.
\newblock Hierarchical text-conditional image generation with {CLIP} latents.
\newblock {\em CoRR}, abs/2204.06125, 2022.

\bibitem{dalle}
Aditya Ramesh, Mikhail Pavlov, Gabriel Goh, Scott Gray, Chelsea Voss, Alec
  Radford, Mark Chen, and Ilya Sutskever.
\newblock Zero-shot text-to-image generation.
\newblock In {\em ICML}, 2021.

\bibitem{Reed2016GenerativeAT}
S. Reed, Zeynep Akata, Xinchen Yan, L. Logeswaran, B. Schiele, and H. Lee.
\newblock Generative adversarial text to image synthesis.
\newblock In {\em ICML}, 2016.

\bibitem{encinstyle}
Elad Richardson, Yuval Alaluf, Or Patashnik, Yotam Nitzan, Yaniv Azar, Stav
  Shapiro, and Daniel Cohen{-}Or.
\newblock Encoding in style: {A} stylegan encoder for image-to-image
  translation.
\newblock In {\em CVPR}, 2021.

\bibitem{interface}
Yujun Shen, Jinjin Gu, Xiaoou Tang, and Bolei Zhou.
\newblock Interpreting the latent space of gans for semantic face editing.
\newblock In {\em CVPR}, 2020.

\bibitem{zhou2021fact}
Yujun Shen and Bolei Zhou.
\newblock Closed-form factorization of latent semantics in gans.
\newblock In {\em CVPR}, 2021.

\bibitem{tewari2020stylerig}
Ayush Tewari, Mohamed Elgharib, Gaurav Bharaj, Florian Bernard, Hans-Peter
  Seidel, Patrick P{\'e}rez, Michael Zollh{\"o}fer, and Christian Theobalt.
\newblock {StyleRig}: Rigging {StyleGAN} for 3d control over portrait images.
\newblock {\em CVPR}, 2020.

\bibitem{bovik}
Z. Wang, E.P. Simoncelli, and A.C. Bovik.
\newblock Multiscale structural similarity for image quality assessment.
\newblock In {\em The Thrity-Seventh Asilomar Conference on Signals, Systems \&
  Computers, 2003}, volume~2, pages 1398--1402 Vol.2, 2003.

\bibitem{stylespace}
Zongze Wu, Dani Lischinski, and Eli Shechtman.
\newblock Stylespace analysis: Disentangled controls for stylegan image
  generation.
\newblock In {\em CVPR}, 2021.

\bibitem{xia2020tedigan}
Weihao Xia, Yujiu Yang, Jing-Hao Xue, and Baoyuan Wu.
\newblock {TediGAN}: Text-guided diverse face image generation and
  manipulation.
\newblock {\em CVPR}, 2021.

\bibitem{leakyrelu}
Bing Xu, Naiyan Wang, Tianqi Chen, and Mu Li.
\newblock Empirical evaluation of rectified activations in convolutional
  network.
\newblock {\em CoRR}, abs/1505.00853, 2015.

\bibitem{Xu2018AttnGANFT}
T. Xu, Pengchuan Zhang, Qiuyuan Huang, Han Zhang, Zhe Gan, Xiaolei Huang, and
  X. He.
\newblock {AttnGAN}: Fine-grained text to image generation with attentional
  generative adversarial networks.
\newblock {\em CVPR}, 2018.

\bibitem{zhang2017stackgan}
Han Zhang, Tao Xu, Hongsheng Li, Shaoting Zhang, Xiaogang Wang, Xiaolei Huang,
  and Dimitris~N Metaxas.
\newblock {StackGAN}: Text to photo-realistic image synthesis with stacked
  generative adversarial networks.
\newblock In {\em ICCV}, 2017.

\bibitem{Zhang2019StackGANRI}
Han Zhang, T. Xu, Hongsheng Li, Shaoting Zhang, Xiaogang Wang, Xiaolei Huang,
  and Dimitris~N. Metaxas.
\newblock {StackGAN++}: Realistic image synthesis with stacked generative
  adversarial networks.
\newblock {\em IEEE TPAMI}, 41:1947--1962, 2019.

\bibitem{lpips}
Richard Zhang, Phillip Isola, Alexei~A. Efros, Eli Shechtman, and Oliver Wang.
\newblock The unreasonable effectiveness of deep features as a perceptual
  metric.
\newblock In {\em 2018 {IEEE} Conference on Computer Vision and Pattern
  Recognition, {CVPR} 2018, Salt Lake City, UT, USA, June 18-22, 2018}, pages
  586--595. Computer Vision Foundation / {IEEE} Computer Society, 2018.

\bibitem{datasetgan}
Yuxuan Zhang, Huan Ling, Jun Gao, Kangxue Yin, Jean{-}Francois Lafleche, Adela
  Barriuso, Antonio Torralba, and Sanja Fidler.
\newblock Datasetgan: Efficient labeled data factory with minimal human effort.
\newblock In {\em CVPR}, 2021.

\bibitem{tigan}
Yufan Zhou, Ruiyi Zhang, Jiuxiang Gu, Chris Tensmeyer, Tong Yu, Changyou Chen,
  Jinhui Xu, and Tong Sun.
\newblock Tigan: Text-based interactive image generation and manipulation.
\newblock In {\em AAAI}, 2022.

\end{thebibliography}
